\pdfoutput=1

\documentclass[11pt]{article}

\usepackage{acl}

\usepackage{times}
\usepackage{latexsym}

\usepackage[T1]{fontenc}

\usepackage[utf8]{inputenc}

\usepackage{microtype}

\usepackage{inconsolata}

\usepackage{graphicx}
\usepackage{cleveref}
\usepackage{xspace}

\usepackage{booktabs}
\usepackage{multirow}

\usepackage{amssymb}
\usepackage{pifont}
\usepackage[framemethod=TikZ]{mdframed}
\usepackage{graphicx}
\usepackage{subcaption}

\newmdenv[%
    backgroundcolor=gray!10,
    linecolor=gray!70,
    linewidth=1pt,
    roundcorner=5pt,
    skipabove=10pt,
    skipbelow=10pt,
    font=\ttfamily\small,
]{promptbox}

\newcommand\blfootnote[1]{%
  \begingroup
  \renewcommand\thefootnote{}\footnote{#1}%
  \addtocounter{footnote}{-1}%
  \endgroup
}

\title{\textsc{RomanSetu}: Efficiently unlocking multilingual capabilities of \\Large Language Models via Romanization}

\author{Jaavid Aktar Husain$^{1,2\dagger\mathsection}$ \quad Raj Dabre$^{1,5,8,9\mathsection}$ \quad Aswanth Kumar$^{3}$ \quad Jay Gala$^{4\ddagger}$ \\ \textbf{Thanmay Jayakumar}$^{1}$ \quad \textbf{Ratish Puduppully}$^6$ \quad \textbf{Anoop Kunchukuttan}$^{1,7,8\mathsection}$ 
  \\[2ex]
  $^{1}$Nilekani Centre at AI4Bharat \quad $^{2}$IIIT D\&M Kancheepuram \quad $^{3}$Flipkart \\
  $^{4}$Mohamed bin Zayed University of Artificial Intelligence \\
  $^{5}$National Institute of Information and Communications Technology, Kyoto, Japan \\
  $^{6}$Institute for Infocomm Research (I$^2$R), A$^{*}$STAR, Singapore \\
  $^{7}$Microsoft, India \quad $^{8}$Indian Institute of Technology Madras, India \\
  $^{9}$Indian Institute of Technology Bombay, India
}

\begin{document}
\maketitle

\blfootnote{$^\dagger$\;Work done during internship at AI4Bharat}
\blfootnote{$\ddagger$\;Work done during employment at AI4Bharat}

\blfootnote{$^\mathsection$\;Corresponding Authors: Jaavid Aktar Husain (\href{mailto:jaavidaktar@gmail.com}{jaavidaktar@gmail.com}), Raj Dabre (\href{mailto:raj.dabre@nict.go.jp}{raj.dabre@nict.go.jp}) and Anoop Kunchukuttan (\href{mailto:ankunchu@microsoft.com}{ankunchu@microsoft.com})}

\begin{abstract}
This study addresses the challenge of extending Large Language Models (LLMs) to non-English languages that use non-Roman scripts. We propose an approach that utilizes the romanized form of text as an interface for LLMs, hypothesizing that its frequent informal use and shared tokens with English enhance cross-lingual alignment. Our approach involves the continual pretraining of an English LLM like Llama 2 \citep{touvron2023llama} on romanized text of non-English, non-Roman script languages, followed by instruction tuning on romanized data. The results indicate that romanized text not only reduces token fertility by 2x-4x but also matches or outperforms native script representation across various NLU, NLG, and MT tasks. Moreover, the embeddings computed on romanized text exhibit closer alignment with their English translations than those from the native script. Our approach presents a promising direction for leveraging the power of English LLMs in languages traditionally underrepresented in NLP. Our code is available on \url{https://github.com/AI4Bharat/romansetu}.
\end{abstract}

\section{Introduction}

Large Language Models (LLMs) demonstrate remarkable proficiency across a broad spectrum of natural language processing (NLP) tasks, as evidenced by various studies \citep{liu2023pre, chung2022scaling, JMLR:v24:22-1144, 10.5555/3600270.3602070, goyal2022news}. They excel not only in tasks for which they were explicitly trained but also in those for which they were not trained. This achievement is mainly due to the availability of corpora \citep{wenzek-etal-2020-ccnet, abadji2021ungoliant, suarez2019asynchronous} as well as the advancements in LLMs that leverage these datasets for pretraining \citep{touvron2023llama, workshop2022bloom, JMLR:v24:22-1144}. Despite their proficiency in English, these models typically demonstrate reduced effectiveness when applied to non-English languages, highlighting a significant challenge in extending their benefits to non-English languages.

\begin{figure*}[t]
    \centering
    \includegraphics[scale=0.3]{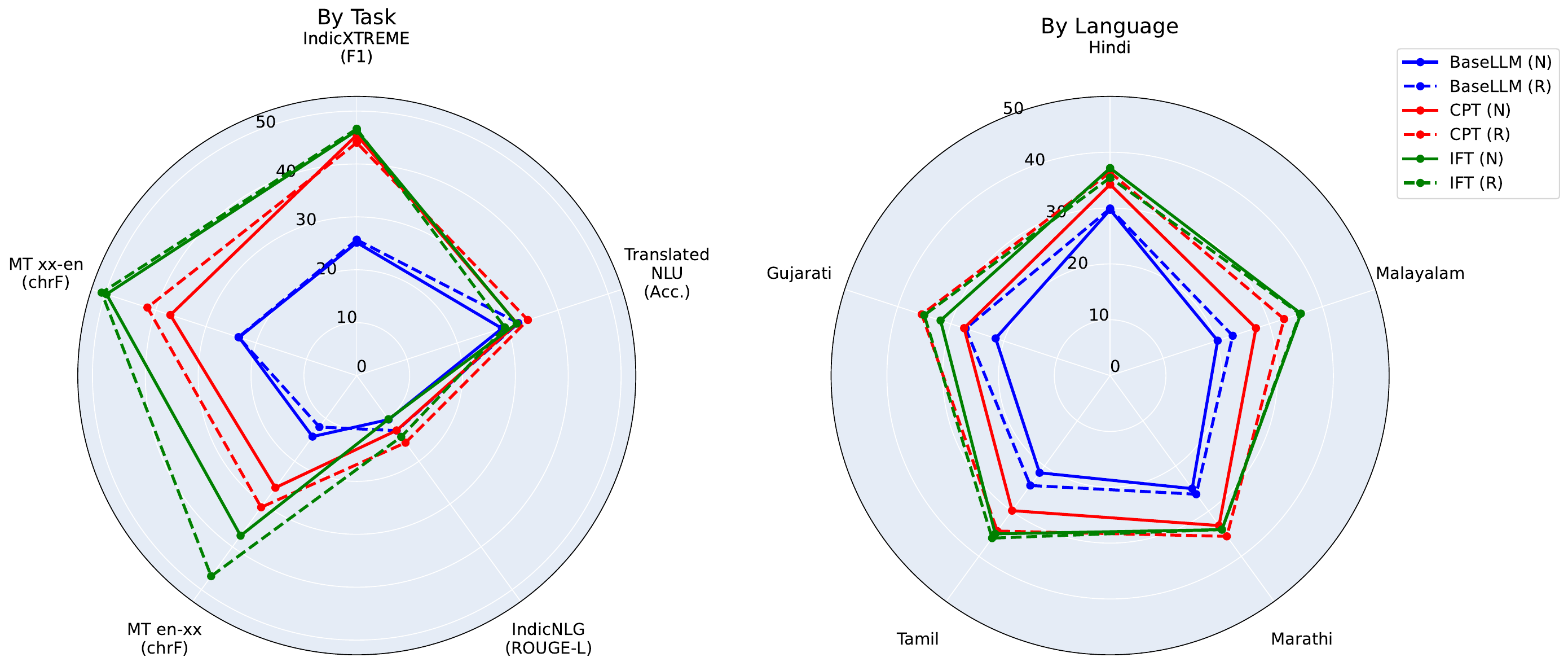}
    \caption{Performance of the BaseLLM (Llama 2 Base), our continually pretrained model (CPT), and our instruction finetuned model (IFT), in both native (N) and romanized (R) script settings. The average scores for different tasks (left) and languages (right) are compared in each radar chart. For CPT, we show 3-shot results where available, or else 1-shot results. For IFT, we use zero-shot results.}
    \label{fig:radarplot}
\end{figure*}

The English-heavy LLMs \citep{touvron2023llama,jiang2023mistral,zhang2022opt} still have some representation coverage from other languages due to data leakage while creating the pre-training dataset, particularly for languages that use the same script as English \textit{i.e.,}  the Roman script. This script sharing enables cross-lingual transfer and bestows some of these LLM capabilities to these languages. For languages using non-Latin scripts, the data representation is very limited to non-existent. Tokenization of text in the languages exhibits high fertility \citep{judit-bert-fertility} and byte-level representation \citep{artetxe-etal-2020-cross} due to inadequate representation of these languages in the tokenizer vocabulary. Hence, these LLMs perform poorly on most of these languages, and inefficient tokenization also leads to high inference latency as well as the inability to process long sequences. This disparity in performance raises a critical question: \textit{how can we extend the capabilities of LLMs to the languages written in non-Latin scripts?}

A widely explored solution is the extension of the tokenizer vocabulary to incorporate new languages and continual pre-training on native language data \citep{cui2023efficient, nguyen2023seallms,minixhofer-etal-2022-wechsel}. This approach is computationally demanding since the models need to be continually pre-trained long enough to effectively learn the new embeddings and align the representations of English and the new language. Furthermore, this approach requires the availability of large volumes of text corpora.

In this work, we explore an alternative approach for more efficient knowledge transfer. Instead of utilizing the native script, we use the romanized form of text as the interface to the LLMs in our approach \textsc{\texttt{RomanSetu}}.\footnote{The word \textit{setu} in Sanskrit means \textit{bridge}, referring to the Roman script serving as a bridge between English and other languages.} The adoption of romanized representation is justified for several reasons. In many languages, romanized text has been frequently used in informal settings and on social media in recent history. This usage creates the potential for the inclusion of some romanized data representation in the pre-training corpus. Additionally, code-mixing with English is a common occurrence and the romanized form shares tokens with English. This leads us to hypothesize that the romanized form is better aligned with English than the native script, thereby facilitating a more effective transfer from English.

In our approach, we first continually pre-train an English LLM like Llama 2 \citep{touvron2023llama} on romanized text for a language generated via transliteration. Subsequently, we perform instruction tuning on this continually pretrained model. We experiment with 5 languages from two language families and evaluate our approach on several NLU, NLG and MT benchmarks. Our experiments and analysis indicate that:

\begin{itemize}
    \item  The fertility of romanized text is 2x-4x smaller than native text, making the romanized form far more efficient than the native script.
    
    \item  The embeddings of the romanized text are closer to the embeddings of their English translations, compared to the embeddings of their native script equivalents, suggesting the former's suitability for cross-lingual transfer.    
    
    \item  Results across several NLU, NLG, and MT tasks show that multilingual instruction finetuning on romanized data produces competitive or superior results than instruction finetuning on native script data, highlighting efficiency and better cross-lingual transfer. \Cref{fig:radarplot} summarizes these results of various models on romanized and native inputs.
    
    \item Specifically, for generation tasks we see significant improvement due to utilization of romanization. To the best of our knowledge, ours is the first work that shows that romanization can help natural language generation tasks. 
    
    \item Romanized representation can enable cross-lingual transfer in decoder-only English-heavy LLMs. Previous work has primarily focused on the use of romanization in multilingual, encoder-only models. 
\end{itemize}


\section{Related Work}

Transliteration refers to the conversion of text written in one script to another. Romanization is a specific instance of transliteration, where the target script is the Roman/Latin script. Romanization has special significance since Roman script is by far the most widely adopted script in the world and many language models are primarily trained for English, which is written in the Roman script. Transliteration is typically used to represent different languages in the same script to enable cross-lingual transfer.

In NMT, \citet{amrhein-sennrich-2020-romanization} show that transliteration shows improvements for low-resource languages with different scripts by transferring from related high-resource languages that use different scripts. \citet{goyal-etal-2020-contact} show that transliteration helps even when only a contact relationship exists between the languages involved. \citet{song-etal-2020-pre} show that transliteration during the pretraining stage for NMT also helps cross-lingual transfer. 

Transliteration has also been used for cross-lingual transfer in the context of pretrained language models. Some works \citep{khemchandani-etal-2021-exploiting,dhamecha-etal-2021-role,moosa-etal-2023-transliteration,purkayastha-etal-2023-romanization} employ transliteration to a common script during the pretraining phase to enable cross-lingual transfer. Other studies \citep{dabre-etal-2022-indicbart,muller-etal-2021-unseen,chau-smith-2021-specializing} adopt transliteration during the fine-tuning.
Transliteration could be done to a common non-Latin script \citep{khemchandani-etal-2021-exploiting,dhamecha-etal-2021-role,dabre-etal-2022-indicbart,doddapaneni-etal-2023-towards} or to the Latin script \cite{muller-etal-2021-unseen,moosa-etal-2023-transliteration,purkayastha-etal-2023-romanization}.  

While transliteration has been explored for language modeling as described above, our work differs from previous work in some important aspects: 

\begin{itemize}
    \item  Previous work explored cross-lingual transfer using transliteration with multilingual language models. We focus our attention on English LLMs and try to achieve cross-lingual transfer via romanization using English LLMs. This is a challenging scenario since the language can be unrelated to English, and very little (if any) native or romanized data in the language under study might be seen during pre-training of the English LLM. At the same time, this is a very practical need since most best performing LLMs are English-heavy \cite{touvron2023llama,jiang2023mistral} and hence cross-lingual transfer via romanization is an important direction to explore. 
    
    \item  We investigate the utility of transliteration in decoder-only language models, which is currently the standard architecture for LLMs. In contrast, all previous work explored cross-lingual transfer with transliteration in the context of encoder-only models (the exception is \citet{dabre-etal-2022-indicbart} which use encoder-decoder models).
    
    \item  While previous work has explored the use of romanization for cross-lingual transfer in language understanding tasks, we show that romanization can be beneficial for generation tasks as well.   
\end{itemize}




To the best of our knowledge, no prior research has investigated leveraging of romanization for mostly English LLMs in the context of cross-lingual transfer to non-English languages.

\begin{figure}[t]
  \includegraphics[scale=0.4]{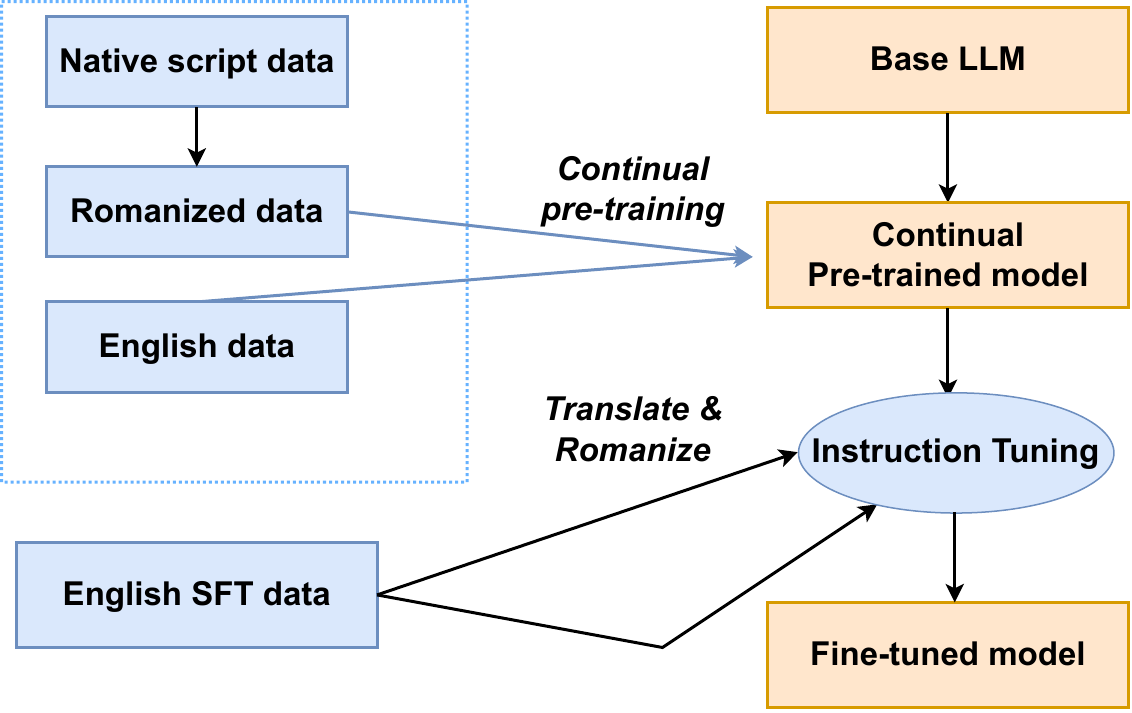}
  \caption{Overview of our proposed approach.}
  \label{fig:flow-diagram}
\end{figure}

\section{Utilizing Romanized Data to make LLMs Multilingual}

Our proposed approach aims to enhance the capabilities of mostly English LLMs for non-English languages by leveraging romanized data. We first continually pretrain the LLM with both romanized and English data to create a base LM that is romanization-aware.  Subsequently, we extend the process by instruction tuning the continually pretrained model. The framework of our approach is illustrated in \Cref{fig:flow-diagram}, encompassing the stages of data romanization, pretraining, and instruction tuning.

\subsection{Romanization Scheme}

A variety of romanization schemes are available, each driven by multiple considerations. One key factor is the resemblance of the romanized representation to the way people typically write romanized text. This is particularly advantageous if the pretraining data includes romanized text, as it might aid in aligning with English. Another important aspect is the fertility achieved by the romanization scheme when considering the original LLM's tokenizer. A third consideration is whether the transliteration scheme is lossy or lossless. A lossless scheme is preferable when the objective is to convert the output back to the native script. Typically, deterministic transliteration schemes are lossless, whereas natural transliteration schemes tend to be lossy.

In this work, we have focused on Indic languages. Hence, we evaluated two romanization schemes for Indic languages: (a) the extended ITRANS scheme from the IndicNLP library \cite{kunchukuttan2020indicnlp}, which defines a fixed, reversible mapping between Indic scripts and Roman characters, and (b) the IndicXlit scheme \cite{madhani-etal-2023-aksharantar}, which generates romanizations as commonly used by Hindi speakers in informal contexts and is learned from parallel transliteration corpora. These mappings are inherently lossy. The IndicXlit romanization demonstrates lower fertility compared to ITRANS romanization (See \Cref{tab:fertility_score}). Preliminary prompting experiments with the Llama 2 model also indicated that IndicXlit outperforms ITRANS in romanized Hindi to English translation. Consequently, we selected IndicXlit as our romanization scheme for this project. However, it is important to note that the transliterations are not reversible. With continued pretraining, ITRANS transliterations might eventually achieve similar task performance to IndicXlit while ensuring script reversibility. This direction might also be relevant for other languages which do not have a transliteration model available, while romanization mappings are available for almost all scripts. We leave this exploration for future work.

\subsection{Continual Pretraining}

While the base model is trained on roman script, most of the data it has been exposed to is in English. Hence, the base model cannot be used as-is for processing romanized inputs, particularly for generation into non-English languages. Hence, the base model needs to be updated. To make the base model romanization-aware, we continue to pretrain the base LLM with romanized document-level data. To prevent any catastrophic forgetting of English capabilities, we also incorporate an equal amount of English data into the pretraining mix. We perform full-finetuning of the models. We do not expand or change the model vocabulary, hence the romanized inputs can take advantage of the token embeddings from the base model while adapting to the new languages.

\subsection{Instruction Finetuning (IFT)}

Non-English languages possess very limited finetuning data. A common method to overcome this limitation is to translate English IFT datasets into the languages. We could choose to translate both input and outputs \cite{wei2023polylm}, or translate only the inputs and ask a powerful, proprietary LLM like ChatGPT to generate translations in the native language \cite{li2023bactrianx}. We chose the former approach since the quality of LLM responses (even for the best commercial models) might not be of good quality \cite{ahuja-etal-2023-mega}, while high-quality open-source translation models \cite{Costa-jussa2024,ai4bharat2023indictrans2} are available making translation of both inputs and outputs feasible at scale for many languages. The translated IFT datasets are then romanized using IndicXlit.



\section{Experimental Settings}

We conducted comprehensive experiments spanning multiple languages, benchmarks and models. This section elaborates on the experimental settings. In all our experiments, we experiment with Llama 2 7B \citep{touvron2023llama} as the base model under various settings. 

\subsection{Languages}
We experimented with the following Indic languages: Hindi, Marathi, Gujarati, Tamil and Malayalam. These languages span 2 language families and 4 different scripts and are mid to low-resource languages. The languages considered in this study belong to the Indo-Aryan branch of Indo-European (Hindi, Marathi, Gujarati) and Dravidian (Tamil, Malayalam) families. In these languages, the typical word order is SOV. These languages have an inflectional morphology, with the Indo-Aryan leaning somewhat towards fusional, while Dravidian languages tend to have an agglutinative morphology. It is important to note that the script systems used for these languages are abugida, effectively faithfully capturing vowels and consonants. We see the choice of Indian languages as a strength since it spans multiple language families, multiple scripts, and diverse linguistic characteristics and resource levels. Past work has also used Indic languages as a case study for romanized representation \cite{moosa-etal-2023-transliteration,dhamecha-etal-2021-role}.

\subsection{Datasets}

\noindent\textbf{Continual Pretraining:} For continual pretraining, we sourced approximately 500 million words of document-level data from web-crawled corpora \citep{doddapaneni-etal-2023-towards} for each language under consideration along with English. To generate the romanized dataset, we transliterated the native script dataset using the \textit{IndicXlit} model \citep{madhani-etal-2023-aksharantar}, a state-of-the-art open-source transliteration model for Indian languages. Both the native script dataset and its romanized counterpart were then used for continual pretraining in various configurations explained later.

\noindent\textbf{Instruction Fine-tuning:} These languages have very little native instruction tuning data for diverse tasks. Following \citet{wei2023polylm}, we rely on translating high-quality English-supervised instruction-tuning datasets into the languages under consideration. We use IndicTrans2 \citep{ai4bharat2023indictrans2}, the state-of-the-art open-source MT model for Indian languages compared to commercial offerings for translation. We sampled examples from various English instruction tuning datasets to ensure a diverse mix of tasks from the Flan collection \citep{longpre2023flan} (65k) and Dolly \citep{DatabricksBlog2023DollyV2} (15k). These are translated into all the languages. The instruction tuning dataset includes a translation subset as well to learn the translation task as well as to drive alignment between English and other languages. We sample 20k high-quality manually translated examples in each direction from the BPCC-Human subset \citep{ai4bharat2023indictrans2}. Thus, the final IFT dataset has 120k examples per language. The instruction tuning datasets are further romanized using IndicXlit.

\noindent\textbf{Evaluation Data:} We evaluate our models on a variety of NLU and NLG tasks. These include native language test sets from the IndicXTREME \citep{doddapaneni-etal-2023-towards} and Indic NLG Suite \citep{kumar-etal-2022-indicnlg} benchmarks. For the translation task, we use the FLORES-200 devtest \citep{goyal-etal-2022-flores,Costa-jussa2024} for evaluation. Following recent work \citep{lai-etal-2023-okapi,üstün2024aya,openai2023gpt4}, we evaluate the knowledge and reasoning capabilities of the model on the machine-translated versions of the benchmarks such as MMLU \citep{hendrycks2021measuring}, ARC \citep{clark2018think}, BoolQ \citep{clark2019boolq} and CommonSenseQA \citep{talmor-etal-2019-commonsenseqa}. Note that as the CommonSenseQA benchmark comprises of a blind test set, we report the evaluations on the dev set. The translations were obtained using IndicTrans2 and the romanized inputs were created using IndicXlit.

\subsection{Training and Finetuning Details}
The models are fully finetuned during continual pre-training as well as supervised fine-tuning. Both the continued pretraining and instruction fine-tuning are multilingual i.e. a single model is trained jointly with data for all languages (including English to preserve English performance). Note that no vocabulary expansion is done. For pre-training, the loss is computed for all the tokens in the sequence. During instruction fine-tuning, the loss was computed only for the output tokens. 
Detailed information about the hyperparameters used for both continual pretraining and instruction fine-tuning can be found in Appendix \ref{apx:hyperparameters}. We adapt \emph{open-instruct}\footnote{\url{https://github.com/allenai/open-instruct/tree/main}} \citep{wang2023far} for our continual pretraining and fine-tuning experiments. 

\subsection{Trained Models}
We begin with Llama 2 as the base model. Then, we create two variants that have undergone continual pre-training: one with native script data (\texttt{CPT-N}) and the other with romanized data (\texttt{CPT-R}). 
For each of these continually pre-trained models, we further perform instruction fine-tuning (IFT). We fine-tune \texttt{CPT-N} using native script IFT data to obtain \texttt{IFT-N}, and \texttt{CPT-R} is fine-tuned with romanized script IFT data to produce \texttt{IFT-R}.

\subsection{Evaluation Metrics}

The evaluation metric for machine translation in our experiments is chrF \citep{papineni2002bleu} computed using the SacreBLEU toolkit \citep{post-2018-call}.\footnote{\texttt{nrefs:1|case:mixed|eff:yes|nc:6|nw:0|space:no|\\version:2.3.1}} For other NLG tasks, we report Rouge-L \cite{lin-2004-rouge} scores. \textit{We primarily report evaluations on the native script outputs. In the case of romanized outputs, we transliterate them back to native script using IndicXlit combined with a unigram word-level language model \cite{madhani-etal-2023-aksharantar}.} Combining word-level results with a unigram language model is efficient and preserves most gains of incorporating a complete n-gram language model \cite{10.1162/coli_a_00510}. In this native script evaluation, the roman script model outputs can also have transliteration errors, which put them at a disadvantage compared to the native script model. To purely compare task performance, we also report results on romanized outputs. For this, we romanize the native script outputs using IndicXlit. Since the final output is an artifact of the same romanization model, this evaluation compares only the task performance.       

\subsection{Models}
We use the 7B-parameter Llama 2 base model as our starting point for all experiments. We evaluate the following models: First, the base Llama 2 model (\texttt{base}) is assessed using both native script and romanized inputs. 
We then evaluate the continually pre-trained models with their respective script data; thus, \texttt{CPT-N} is evaluated with native script data, and \texttt{CPT-R} is evaluated with romanized data. Additionally, the instruction-tuned models are evaluated with their respective script data; therefore, \texttt{IFT-N} is assessed with native script data, and \texttt{IFT-R} is assessed with romanized data.

\subsection{Prompting and Decoding}

For the base and CPT models, we do 1-shot and 3-shot prompting. For IFT models, we experiment with 0-shot prompting. 
The prompt templates for various tasks are shown in \Cref{apx:prompts}. 
For generation tasks, we use greedy decoding. For classification tasks, we completed the prompts with various possible options as per class labels and computed the output probabilities. We choose the class that maximizes the probability \citep{liu2021pretrain}.

\section{Results}

In this section, we present the results of various experiments comparing romanized and native representations and discuss the results. We aim to address the following research questions:
\begin{itemize}
    \item Is there an efficiency gain through romanization, and if so, which approach to romanization is the most efficient?
    \item Do romanized inputs lead to embeddings that are better aligned to English representations compared to native inputs?
    \item Does romanization improve downstream task performance across multiple NLP tasks?
\end{itemize}

\begin{table}[t]
\centering
\small
\resizebox{0.8\linewidth}{!}{
\begin{tabular}{lccc}
\toprule
\textbf{Language} & \textbf{N} & \textbf{R} & \textbf{R-IndicNLP} \\
\midrule
Gujarati & 18.44 & \textbf{3.39} & 4.16 \\
Hindi & 7.36 & \textbf{2.98} & 3.98 \\
Malayalam & 12.85 & \textbf{5.04} & 5.56 \\
Marathi & 8.91 & \textbf{3.64} & 4.84 \\
Tamil & 12.11 & \textbf{4.89} & 5.35 \\
\bottomrule
\end{tabular}
}
\caption{Fertility scores on the FLORES-200 dev set \citep{goyal-etal-2022-flores,Costa-jussa2024} for native (N), romanized with IndicXLIT (R) and romanized with IndicNLP library (R-IndicNLP). Bold indicates lowest fertility.}
\label{tab:fertility_score}
\end{table}

\paragraph{Efficiency Gains with Romanization:}
To address this question, we compute fertility scores on the FLORES-200 dev set \citep{goyal-etal-2022-flores,Costa-jussa2024}. Fertility refers to the number of tokens per word generated by the Llama 2 tokenizer \cite{touvron2023llama}. We compare the native script against two forms of romanized text: transliteration using IndicXlit \cite{madhani-etal-2023-aksharantar} and rule-based extended ITRANS transliteration scheme implemented in the IndicNLP library \cite{kunchukuttan2020indicnlp}.
Table \ref{tab:fertility_score} presents the fertility scores on the Llama 2 model for the languages we experimented with. We observed that the fertility (and hence sequence length) in the native script is at least twice that of the IndicXlit romanized text. In languages like Gujarati, the native script fertility is very large (greater than 4 times the romanized fertility). Hence, we can expect significant efficiency gains when processing romanized text over native text in terms of reduced memory consumption, reduced generation time and increased maximum sequence length limit. 

Furthermore, romanization using IndicNLP \cite{kunchukuttan2020indicnlp} has slightly higher fertility compared to IndicXlit. While we have used IndicXlit-based transliterations in this work for reasons explained earlier, it would be interesting to explore deterministic and reversible romanizations like IndicNLP at the cost of slightly increased inefficiency. The advantage of reversible mapping would be avoiding transliteration errors when mapping the outputs back to the native script. 

\begin{table}[t]
\centering
\small
\resizebox{0.75\linewidth}{!}{
\begin{tabular}{lccc}
\toprule
\textbf{Language} & \textbf{E - N} & \textbf{E - R} & \textbf{N - R} \\
\midrule
Gujarati & 0.39 & 0.47 & 0.51 \\
Hindi & 0.40 & 0.50 & 0.34 \\
Malayalam & 0.40 & 0.46 & 0.52 \\ 
Marathi & 0.44 & 0.48 & 0.58 \\ 
Tamil & 0.44 & 0.43 & 0.53 \\
\bottomrule
\end{tabular}
}
\caption{Cosine similarities are computed using the last token representations of the last layer from Llama 2. In this context, E - N represents the cosine similarity between English and the native script, E - R represents the similarity between English and the romanized script and N - R represents the similarity between the native and the romanized script.}
\label{tab:cosine_sim}
\end{table}

\paragraph{Cross-lingual Representation Similarity:}
We compare the sentence embeddings generated by the native and romanized inputs as well as English on the base model. Specifically, we compare cosine similarities between sentence embeddings generated by these inputs. Sentence embeddings are computed using the last token representation of the final hidden layer. In \Cref{tab:cosine_sim}, we see that the Roman script representations are closer to English representations compared to the native script representations, making them a better candidate for cross-lingual transfer from English.

\paragraph{Machine Translation:}





We discuss the results of machine translation, a canonical multilingual language generation task. 
\Cref{tab:nlg-results} (Rows XX-En translation and En-XX translation) shows results on machine translation both in and out of English.

For translation into English, the base model translation quality is roughly equivalent for both native and roman scripts. We see that continual pretraining helps improve translation quality significantly for both representations. Romanized scripts show significantly more improvement on the CPT models compared to native scripts. We hypothesize that romanization is better able to align cross-lingual representations between English and other languages to improve translation performance. Romanized IFT models also perform better than their native counterparts, albeit the gap is now narrower compared to CPT models.

For translation out of English, the base model performs very poorly which is understandable since the model might have seen only negligible non-English data. CPT significantly improves translation quality on both representations, with romanized representation showing better performance. The \texttt{IFT-R} models are able to take good advantage of the romanized representations, and they show an average improvement of 8 chrF points over the corresponding native script \texttt{IFT-N} models. Thus, we show that even generation into romanized script can take advantage of the English capabilities of the LLM. 

\Cref{tab:translation_results} shows the effect of the the choice of script when evaluating translation out of English to each language. 
We also report evaluation on Roman script outputs for both models and see that the romanized IFT models report even larger gains in this evaluation setting. This suggests that the true gains from romanization might be larger and it is not reflected in native script evaluation on account of errors in Roman to native transliteration that is performed for the evaluation.

\begin{table}[]
\centering
\small
\resizebox{\linewidth}{!}{
\begin{tabular}{lcccccc}
\toprule
\multirow{2}{*}{} & \multirow{2}{*}{\textbf{Script}} & \multicolumn{2}{c}{\textbf{BaseLLM}} & \multicolumn{2}{c}{\textbf{CPT}} & \textbf{IFT} \\
\cmidrule{3-7}
&   & \textbf{1-shot} & \textbf{3-shot} & \textbf{1-shot} & \textbf{3-shot} & \textbf{0-shot} \\
\midrule
\multirow{2}{*}{XX-En translation}   & N & 22.46           & 23.42           & 38.54           & 37.07           & 49.78           \\
   & R & \textbf{22.52}  & \textbf{23.52}  & \textbf{42.53}  & \textbf{41.64}  & \textbf{50.75}  \\
\midrule
\multirow{2}{*}{En-XX translation} & N & 13.95           & \textbf{14.25}  & 25.55           & 26.19           & 37.40           \\
   & R & \textbf{14.20}  & 12.02           & \textbf{29.55}  & \textbf{30.77}  & \textbf{46.87}  \\
\midrule
\multirow{2}{*}{XLSum}                 & N & 6.88            & -               & 7.59            & -               & 7.77            \\
   & R & \textbf{10.16}   & -               & \textbf{12.44}   & -               & \textbf{12.56}   \\
\midrule
\multirow{2}{*}{IndicHeadline}         & N & 13.66           & -               & 18.04           & -               & 12.61           \\
   & R & \textbf{15.56}  & -               & \textbf{18.92}  & -               & \textbf{16.03} \\
\bottomrule
\end{tabular}
}
\caption{Results on NLG Benchmarks - averaged across languages (RougeL score for XLSum and IndicHeadline; chrF for MT. En-XX refers to out-of-English and XX-En refers to into-English directions. N and R indicate whether the model was trained and decoded on native or romanized script. Evaluation is always in the native script.}
\label{tab:nlg-results}
\end{table}

\begin{table}[t]
\small
\centering
\resizebox{0.95\linewidth}{!}{
\begin{tabular}{lcccc}
\toprule
\multirow{2}{*}{\textbf{Language}} & \multicolumn{2}{c}{\textbf{Native-Script Eval}} & \multicolumn{2}{c}{\textbf{Roman-Script Eval}} \\
\cmidrule{2-5}
 & \textbf{N} & \textbf{R} & \textbf{N} & \textbf{R} \\
\midrule
Gujarati & 23.9 & \textbf{26.0} & 45.4 & \textbf{53.5} \\
Hindi & 44.1 & \textbf{49.1} & 45.9 & \textbf{56.2} \\
Malayalam & 39.4 & \textbf{42.8} & 49.0 & \textbf{53.8} \\
Marathi & 40.3 & \textbf{44.8} & 44.0 & \textbf{50.7} \\
Tamil & 39.3 & \textbf{41.5} & 50.1 & \textbf{53.7} \\
\bottomrule
\end{tabular}
}
\caption{Native and Roman script evaluation results via chrF scores for En-XX translation on IFT model in 0-shot setting. N and R indicate whether the model was trained and decoded on native or romanized script.}
\label{tab:translation_results}
\end{table}

\paragraph{Language Generation Tasks:} \Cref{tab:nlg-results} shows the results on other generation tasks \textit{viz.} summarization and headline generation. Similar to MT, we observe that models utilizing romanized script consistently outperform those using native script by \detokenize{~}4 RougeL points across both continually pretraining and instruction fine-tuning setups with different shots. On summarization using the XLSum dataset \citep{hasan-etal-2021-xl}, we find that CPT and IFT models demonstrate competitive performance in both native and romanized scripts. A bit surprisingly, we find that IFT models underperform by 3-5 RougeL points compared to CPT models on IndicHeadline. We hypothesize that the format of the example was not seen in the IFT dataset, resulting in poorer zero-shot performance compared to the one-shot performance of the CPT model. Overall, we find romanization to benefit not only in terms of efficiency but also in terms of downstream performance.

\begin{table}[t]
\centering
\small
\resizebox{\linewidth}{!}{
\begin{tabular}{lcccccc}
\toprule
\multirow{2}{*}{} & \multirow{2}{*}{\textbf{Script}} & \multicolumn{2}{c}{\textbf{BaseLLM}} & \multicolumn{2}{c}{\textbf{CPT}} & \textbf{IFT} \\
\cmidrule{3-7}
&   & \textbf{1-shot} & \textbf{3-shot} & \textbf{1-shot} & \textbf{3-shot} & \textbf{0-shot} \\
\midrule
\multirow{2}{*}{IndicSentiment} & N                       & 45             & \textbf{44.32} & 70.58          & 87.32       & \textbf{88.88} \\
   & R & \textbf{48.16} & 43.7           & \textbf{88.46} & \textbf{92.82} & 82.44   \\
\midrule
\multirow{2}{*}{IndicCOPA}      & N                       & \textbf{21.08} & \textbf{25.43} & \textbf{20.47} & \textbf{35} & 37.3           \\
   & R & 17.53          & 8.12           & 18.23          & 18.78          & \textbf{45.45} \\
\midrule
\multirow{2}{*}{IndicXNLI} & N & 16.98          & 21.1           & 18.68          & 35.76          & \textbf{42.3}  \\
   & R & \textbf{27.24} & \textbf{31.16} & \textbf{36.8}  & \textbf{37.28} & 38.38 \\
\midrule
\multirow{2}{*}{\begin{tabular}[c]{@{}l@{}}IndicQA\\ (with context)\end{tabular}} & N                       & 9.77           & -              & 23.83          & -   & 16.79 $^\dagger$          \\
  & R  & \textbf{19.74} & -    & \textbf{27.25} & -              & \textbf{20.33} $^\dagger$ \\
\bottomrule
\end{tabular}
}
\caption{Results on IndicXTREME NLU Testsets (average F1 score across languages). $^\dagger$ indicates 1-shot as scores for 0-shot were close to zero and non-interpretable so we made an exception for this task. N and R indicate whether the model was trained and decoded on native or romanized script. Evaluation is always in the native script.}
\label{tab:indicxtreme-nlu-results}
\end{table}

\begin{table}[t]
\centering
\small
\resizebox{\linewidth}{!}{
\begin{tabular}{lcccccc}
\toprule
\multirow{2}{*}{}              & \multirow{2}{*}{\textbf{Script}} & \multicolumn{2}{c}{\textbf{BaseLLM}} & \multicolumn{2}{c}{\textbf{CPT}} & \textbf{IFT}   \\
    \cmidrule{3-7}
   &   & \textbf{1-shot} & \textbf{3-shot} & \textbf{1-shot} & \textbf{3-shot} & \textbf{0-shot} \\
\midrule
\multirow{2}{*}{MMLU}                      & N & 26.87           & 27.43           & 28.70           & 28.98           & 24.59           \\
   & R & \textbf{26.94}  & \textbf{27.48}  & \textbf{30.59}  & \textbf{31.14}  & \textbf{26.28}  \\
\midrule
\multirow{2}{*}{BoolQ}                     & N & 57.90           & 44.45  & 59.09           & 47.48           & \textbf{60.04}  \\
   & R & \textbf{60.75}  & \textbf{61.03}  & \textbf{61.85}  & \textbf{57.10}  & 46.95  \\
\midrule
\multirow{2}{*}{Arc Easy}                  & N & 25.82           & 26.35           & \textbf{28.77}  & \textbf{28.25}  & \textbf{26.01}  \\
   & R & \textbf{26.38}  & \textbf{26.44}  & 26.69  & 27.10           & 25.34  \\
\midrule
\multirow{2}{*}{Arc Challenge} & N                                & \textbf{26.54}    & \textbf{25.70}   & \textbf{27.56}  & \textbf{27.20} & \textbf{24.20} \\
   & R & 24.61  & 25.56           & 26.40  & 26.66           & 22.80  \\
\midrule
\multirow{2}{*}{CommonSenseQA} & N                                & \textbf{21.01}    & \textbf{20.45}   & \textbf{26.55}  & 27.42          & 24.16          \\
   & R & 20.07           & 20.18           & 25.55           & \textbf{28.18}  & \textbf{25.95}  \\
\midrule
\multirow{2}{*}{\begin{tabular}[c]{@{}l@{}}IndicQA\\ (without context)\end{tabular}} & N & 1.48 & - & 5.56 & - & 1.13 $^\dagger$ \\
 & R & \textbf{12.96} & - & \textbf{17.62} & - & \textbf{16.87} $^\dagger$ \\
 \bottomrule
\end{tabular}
}
\caption{Results on Translated NLU Testsets (average Accuracy score across languages with the exception of IndicQA where use F1 score). $^\dagger$ indicates scores reported for 1-shot as scores for 0-shot were close to zero and non-interpretable so we made an exception for this task. N and R indicate whether the model was trained and decoded on native or romanized script. Evaluation is always in the native script.}
\label{tab:translated-nlu-results}
\end{table}

\paragraph{Language Understanding Tasks:} \Cref{tab:indicxtreme-nlu-results} shows the results on different NLU tasks part of IndicXTREME benchmark \cite{doddapaneni-etal-2023-towards}, including IndicSentiment, IndicCOPA, and IndicXNLI. We find that romanized script models are competitive to those that utilize native script. Specifically, in tasks such as IndicSentiment and IndicXNLI, romanized models consistently outperform their native counterparts across various shots. However, native models generally perform superior compared to romanized variants for the IndicCOPA task, except for the IFT model. Furthermore, we observe notable performance improvements for IndicQA without context (a reading comprehension task) across all models with romanized script when we provide the relevant passage in the context. This indicates that models with the romanized script tend to benefit from knowledge transfer from English and are also efficient in terms of sequence length. 

\paragraph{Knowledge and Reasoning tasks:} \Cref{tab:translated-nlu-results} shows the results on translated versions of various knowledge and reasoning benchmarks such as MMLU \citep{hendryckstest2021}, BoolQ \citep{clark-etal-2019-boolq}, ARC \citep{allenai:arc} and CommonsenseQA \citep{talmor-etal-2019-commonsenseqa}. For most of the tasks, the results of the romanized script are competitive with the native script. At the same time, they are efficient to process, as discussed earlier. For IndicQA without context (open domain question answering task), we observe that the romanized model shows \detokenize{~}10 F1 point gain over the native script model. This demonstrates the overall effectiveness of utilizing romanization for non-Latin script languages.

\paragraph{Discussion:} From the analysis and results presented earlier, we can summarize the following. Romanized representations are significantly more efficient compared to native script representations. They are better aligned to English representations compared to native script representations, hence they are better for cross-lingual transfer. Across a range of tasks and languages, romanized representations provide competitive or better performance over their native script counterparts (detailed results per language and task are presented in Appendix \ref{apx:results}. Specifically, for a generation task like machine translation, we see significant gains while translating out of English with romanized representations. In Appendix \ref{apx:translations}, we show some examples of romanized and native script translations, illustrating the gains from romanization. These results indicate that romanization is a promising alternative to extending the capabilities of English LLMs to other languages using non-Roman scripts.



We primarily base our study on English-only models due to their state-of-the-art performance across a diverse array of tasks and consistent improvements in capabilities.  In contrast, multilingual models like BLOOM significantly lag behind English-only models, even on English tasks, and their progress has been slower. This disparity is due to the limited amount of English data and the lack of fine-tuning data for non-English languages used in training multilingual models. Therefore, improving English-only models is a practical way to enhance multilingual LLMs. We fine-tuned the BLOOM 7B model with the instruction dataset in both native and roman script. We observe that the roman script model outperforms the native script model, consistent with the findings on the Llama 2 base model. We provide the evaluation results for BLOOM 7B model across different tasks in \Cref{apx:bloom-results}.

\section{Conclusion}

In this study, we proposed the use of romanization to extend the performance of LLMs primarily trained in English to other languages. Our approach successfully unlocks LLM capabilities for non-English languages by using romanization to bridge English and non-English language representations. We have empirically demonstrated the effectiveness of this strategy through experiments involving few-shot prompting, continual pretraining, and instruction fine-tuning on a variety of tasks. Additionally, using romanized data increases inference speed and maximum processable sequence length, while reducing memory requirements by 2x to 4x depending on the language.

\section{Future Work}

Looking forward, we also plan to explore reversible, deterministic transliteration so that transliteration errors in output post-processing are eliminated. In addition, it would be interesting to explore if romanized representation can act as a bridge to efficiently improve native script performance also when English LLMs are extended to incorporate non-Roman tokens.

\section{Limitations}

In this work, we chose to use natural romanizations since they tended to have lower fertility and might have been seen in the base LM pretraining. Transliterating the outputs back to the native script can be lossy. We use a high-quality transliteration model to minimize the transliteration errors. However, such high-quality transliterations might not be available for all languages. Using a deterministic and reversible transliteration scheme can alleviate this problem. 

The experiments in this work have been conducted on Indian languages. Indian languages are representative of the scenarios we study and we cover multiple language families and scripts with varying resource status. Previous studies have also used Indic languages as a case study for multilinguality and multi-script scenarios \cite{moosa-etal-2023-transliteration,khemchandani-etal-2021-exploiting}. Hence, we believe the findings should generalize to other language families - but further experimentation can confirm this hypothesis. 


We have done only a limited amount of pretraining data in our experiments due to compute constraints. Resource constraints led us to limit our experiments to a 7B  model, but experiments with larger models can yield more insights. Larger models and pre-training on larger datasets might benefit the multilingual models \cite{kaplan2020scaling,pmlr-v202-fernandes23a}. We expect the broad conclusions to be the same - that romanized models are more efficient and better/comparable in task performance to native-script models.

Some of the evaluations have been done on translated benchmarks. While this is not ideal, most low-resource languages lack testsets across diverse tasks. Hence, multiple works have relied on translated benchmarks to measure multilingual performance - these can give indicative trends on performance across languages. More efforts are needed to create multilingual benchmarks for diverse tasks.

\section{Ethics Statement}

We use romanization as a way to align English with other languages using non-Latin scripts. Given the current state of LLMs, this seems like a practical direction to extend the capabilities of the best LLMs to other languages. The intention is not to supplant the use of a native script (which is widely adopted and has a rich literary tradition) with romanized script but to use romanization as a way to efficiently bring the benefits of LLM technology to low-resource languages written in non-Latin scripts. Further advancements are needed to extend this line of research to improve native script performance efficiently. 

This work does not involve any new data collection and does not employ any annotators for data collection. We utilize publicly available datasets for the experiments reported in this work. Some of these datasets originate from web crawls, and we do not explicitly attempt to identify any biases within these datasets, using them in their original form.

\subsubsection*{Acknowledgments}
We would like to thank EkStep Foundation and Nilekani Philanthropies for their generous grant towards research and building datasets, models, tools and other resources for Indian languages.
\bibliography{anthology,custom}

\appendix
\onecolumn

\section{Model Training Details}
\label{apx:hyperparameters}

Tables \ref{tab:cpt_hyperparameters} and \ref{tab:IFT_hyperparameters} show the hyperparameters used for training the CPT and IFT models. 

\begin{table}[h]
\small
\centering
\begin{tabular}{ll}
\toprule
\textbf{Hyperparameter} & \textbf{Value} \\ 
\midrule
Batch Size (tokens) & 1M \\ 
Learning Rate & 5e-5 \\ 
Number of Epochs & 1 \\ 
Maximum Sequence Length & 2,048 \\
\bottomrule
\end{tabular}
\caption{The range of hyperparameters used for continual pretraining.}
\label{tab:cpt_hyperparameters}
\end{table}

\begin{table}[h]
\small
\centering
\begin{tabular}{ll}
\toprule
\textbf{Hyperparameter} & \textbf{Value} \\ 
\midrule
Batch Size (examples) & 128 \\ 
Learning Rate & 5e-5 \\ 
Number of Epochs & 1 \\ 
Maximum Sequence Length & 2,048\footnotemark \\
\bottomrule
\end{tabular}
\caption{The range of hyperparameters used for supervised fine-tuning.}
\label{tab:IFT_hyperparameters}
\end{table}

\footnotetext{Although this is the maximum permissible length, most examples fall far below this length during fine-tuning.}

\section{Prompt Templates}
\label{apx:prompts}

We list the various prompt templates for evaluations across different test sets (see \Cref{fig:indicsentiment-prompt,fig:indiccopa-prompt,fig:indicxnli-prompt,fig:indicqa-prompt,fig:xlsum-prompt,fig:indicheadline-prompt,fig:mmlu-prompt,fig:boolq-prompt,fig:arc-prompt,fig:commonsenseqa-prompt}).

\section{Translation Examples}
\label{apx:translations}

\begin{figure*}[h]
  \fbox{\includegraphics[width=0.95\linewidth]{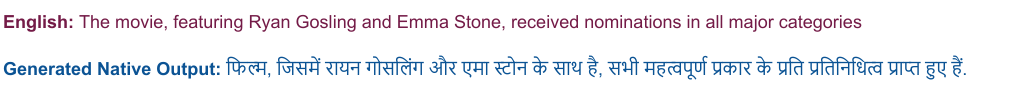}}
  \caption{Example of \textit{zero-shot} translation for English to Hindi machine translation task to generate text in native script using a model fine-tuned on native script data.}
  \label{fig:en_to_xx_native}
\end{figure*}

\begin{figure*}[h]
  \fbox{\includegraphics[width=0.95\linewidth]{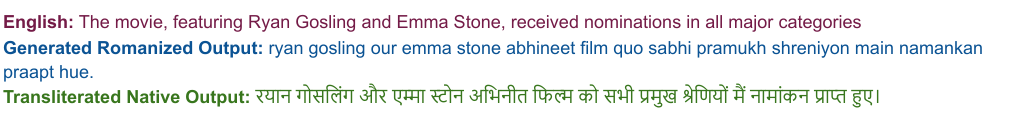}}
  \caption{Example of \textit{zero-shot} translation for English to Hindi machine translation task to generate text in romanized script using a model fine-tuned on romanized data. The romanized output is then transliterated to native script.}
  \label{fig:en_to_xx_native_roman}
\end{figure*}

\begin{figure*}[h]
  \fbox{\includegraphics[width=0.95\linewidth]{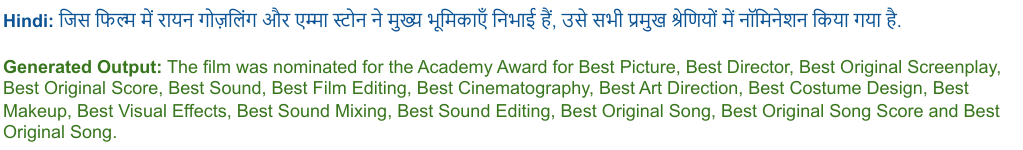}}
  \caption{Example of \textit{zero-shot} translation for the Hindi to English machine translation task using native script input for a model fine-tuned on native script data.}
  \label{fig:xx_to_en_native}
\end{figure*}

\begin{figure*}[h]
  \fbox{\includegraphics[width=0.95\linewidth]{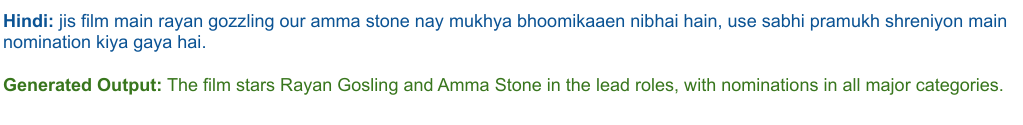}}
  \caption{Example of \textit{zero-shot} translation for the Hindi to English machine translation task using romanized script input for a model fine-tuned on romanized data.}
   \label{fig:xx_to_en_romanized}
\end{figure*}

This section shows some examples of translations using the romanized and native-script IFT models (See \Cref{fig:en_to_xx_native,fig:en_to_xx_native_roman,fig:xx_to_en_native,fig:xx_to_en_romanized}). We can see in these selected examples that romanized representation performs better.

\section{Language Specific Results}
\label{apx:results}

We report the language-wise evaluation results for 5 Indic languages considered for this study across a diverse array of tasks ranging from text generation, text understanding, and text reasoning in \Cref{langwise-flores-enxx,langwise-flores-xxen,langwise-xlsum,langwise-indicheadline,langwise-mmlu,langwise-boolq,langwise-arc-easy,langwise-arc-challenge,langwise-commonsenseqa,langwise-indicsentiment,langwise-indiccopa,langwise-indicxnli,langwise-indicqa-with-context,langwise-indicqa-without-context}.

\section{Results on BLOOM 7B}\label{apx:bloom-results} 

We report evaluation results for 5 Indic languages for the tasks such as machine translation, headline generation and summarization on instruction fine-tuned BLOOM 7B in \Cref{tab:bloom-mt,tab:bloom-indicheadline,tab:bloom-xlsm}.

\newpage

\begin{figure}[h]
\centering
\begin{promptbox}
Predict the sentiment of the review. The possible choices for the sentiment are: 'positive' and 'negative'.\\ \\
Review: \{\{\ text\ \}\} \\
Sentiment: \{\{\ label\ \}\} \\ \\
$\cdots$ \\
$\cdots$ \\ \\
Review: \{\{\ text\ \}\} \\
Sentiment:
\end{promptbox}
\caption{Prompt template for IndicSentiment}
\label{fig:indicsentiment-prompt}
\end{figure}

\begin{figure}[h]
\centering
\begin{promptbox}
I am hesitating between two options. Help me choose the more likely cause or effect.\\ \\
\{\{\ premise\ \}\} \{\{\ connector\ \}\} \\
A. \{\{\ choice1\ \}\} \\
B. \{\{\ choice2\ \}\} \\
Answer: \{\{\ label\ \}\} \\ \\
$\cdots$ \\
$\cdots$ \\ \\
\{\{\ premise\ \}\} \{\{\ connector\ \}\} \\
A. \{\{\ choice1\ \}\} \\
B. \{\{\ choice2\ \}\} \\
Answer: 
\end{promptbox}
\caption{Prompt template for IndicCOPA}
\label{fig:indiccopa-prompt}
\end{figure}

\begin{figure}[!ht]
\centering
\begin{promptbox}
Answer whether the hypothesis is more likely to be true (entailment), false (contradiction), or unknown (neutral) based on the given premise.\\ \\
Premise: \{\{\ premise\ \}\} \\
Hypothesis: \{\{\ hypothesis\ \}\} \\
Answer: \{\{\ label\ \}\} \\ \\
$\cdots$ \\
$\cdots$ \\ \\
Premise: \{\{\ premise\ \}\} \\
Hypothesis: \{\{\ hypothesis\ \}\} \\
Answer: 
\end{promptbox}
\caption{Prompt template for IndicXNLI}
\label{fig:indicxnli-prompt}
\end{figure}

\begin{figure}[!ht]
\centering
\begin{promptbox}
Answer the following question based on the information in the given passage. \\ \\
Passage: \{\{\ passage\ \}\} \\
Question: \{\{\ question\ \}\} \\
Answer: \{\{\ answer\ \}\} \\ \\
$\cdots$ \\
$\cdots$ \\ \\
Passage: \{\{\ passage\ \}\} \\
Question: \{\{\ question\ \}\} \\
Answer: 
\end{promptbox}
\caption{Prompt template for IndicQA.}
\label{fig:indicqa-prompt}
\end{figure}

\begin{figure}[t]
\centering
\begin{promptbox}
Summarize the following \{\{\ language\ \}\} article(s) as accurately as possible in few sentences.\\ \\
\{\{\ language\ \}\} article: \{\{\ article\ \}\} \\
\{\{\ language\ \}\} summary: \{\{\ summary\ \}\} \\ \\
$\cdots$ \\
$\cdots$ \\ \\
\{\{\ language\ \}\} article: \{\{\ article\ \}\} \\
\{\{\ language\ \}\} summary: 
\end{promptbox}
\caption{Prompt template for XLSum.}
\label{fig:xlsum-prompt}
\end{figure}

\begin{figure}[t]
\centering
\begin{promptbox}
Generate a headline for the following article(s) as accurately as possible.\\ \\
\{\{\ language\ \}\} article: \{\{\ article\ \}\} \\
\{\{\ language\ \}\} headline: \{\{\ headline\ \}\} \\ \\
$\cdots$ \\
$\cdots$ \\ \\
\{\{\ language\ \}\} article: \{\{\ article\ \}\} \\
\{\{\ language\ \}\} headline: 
\end{promptbox}
\caption{Prompt template for IndicHeadline.}
\label{fig:indicheadline-prompt}
\end{figure}

\begin{figure}[t]
\centering
\begin{promptbox}
The following are multiple choice questions (with answers) about \{\{\ subject\ \}\}.\\ \\
\{\{\ question\ \}\} \\
A. \{\{\ choice1\ \}\} \\
B. \{\{\ choice2\ \}\} \\
C. \{\{\ choice3\ \}\} \\
D. \{\{\ choice4\ \}\} \\ 
Answer: \{\{\ label\ \}\} \\ \\
$\cdots$ \\
$\cdots$ \\ \\
\{\ subject\ \}\}\\ \\
\{\{\ question\ \}\} \\
A. \{\{\ choice1\ \}\} \\
B. \{\{\ choice2\ \}\} \\
C. \{\{\ choice3\ \}\} \\
D. \{\{\ choice4\ \}\} \\
Answer: 
\end{promptbox}
\caption{Prompt template for MMLU.}
\label{fig:mmlu-prompt}
\end{figure}

\begin{figure}[t]
\centering
\begin{promptbox}
The following are binary yes/no choice questions (with answers).\\ \\
Passage: \{\{\ passage\ \}\} \\
Question: \{\{\ question\ \}\} \\
Answer: \{\{\ label\ \}\} \\ \\
$\cdots$ \\
$\cdots$ \\ \\
Passage: \{\{\ passage\ \}\} \\
Question: \{\{\ question\ \}\} \\
Answer: 
\end{promptbox}
\caption{Prompt template for BoolQ.}
\label{fig:boolq-prompt}
\end{figure}

\begin{figure}[t]
\centering
\begin{promptbox}
The following are multiple choice questions (with answers) about science.\\ \\
\{\{\ question\ \}\} \\
A. \{\{\ choice1\ \}\} \\
B. \{\{\ choice2\ \}\} \\
C. \{\{\ choice3\ \}\} \\
D. \{\{\ choice4\ \}\} \\ 
E. \{\{\ choice5\ \}\} \\ 
Answer: \{\{\ label\ \}\} \\ \\
$\cdots$ \\
$\cdots$ \\ \\
\{\{\ question\ \}\} \\
A. \{\{\ choice1\ \}\} \\
B. \{\{\ choice2\ \}\} \\
C. \{\{\ choice3\ \}\} \\
D. \{\{\ choice4\ \}\} \\ 
E. \{\{\ choice5\ \}\} \\ 
Answer: \{\{\ label\ \}\}
\end{promptbox}
\caption{Prompt template for ARC.}
\label{fig:arc-prompt}
\end{figure}

\begin{figure}[t]
\centering
\begin{promptbox}
The following are multiple choice questions (with answers) requiring common sense.\\ \\
\{\{\ question\ \}\} \\
A. \{\{\ choice1\ \}\} \\
B. \{\{\ choice2\ \}\} \\
C. \{\{\ choice3\ \}\} \\
D. \{\{\ choice4\ \}\} \\ 
E. \{\{\ choice5\ \}\} \\ 
Answer: \{\{\ label\ \}\} \\ \\
$\cdots$ \\
$\cdots$ \\ \\
\{\{\ question\ \}\} \\
A. \{\{\ choice1\ \}\} \\
B. \{\{\ choice2\ \}\} \\
C. \{\{\ choice3\ \}\} \\
D. \{\{\ choice4\ \}\} \\ 
E. \{\{\ choice5\ \}\} \\ 
Answer: \{\{\ label\ \}\}
\end{promptbox}
\caption{Prompt template for CommonsenseQA.}
\label{fig:commonsenseqa-prompt}
\end{figure}

\clearpage

\begin{table*}[]
\small
\centering
\resizebox{\linewidth}{!}{
\begin{tabular}{lcccccccccccc}
\toprule
\multirow{2}{*}{\textbf{Model}} & \multirow{2}{*}{\textbf{Shots}} & \multirow{2}{*}{\textbf{Input text}} & \multicolumn{2}{c}{\textbf{Hindi}} & \multicolumn{2}{c}{\textbf{Gujarati}} &  \multicolumn{2}{c}{\textbf{Tamil}} & \multicolumn{2}{c}{\textbf{Marathi}} & \multicolumn{2}{c}{\textbf{Malayalam}} \\
\cmidrule{4-13}
& & & \multicolumn{1}{c}{\textbf{N}}     & \multicolumn{1}{c}{\textbf{R}}     & \multicolumn{1}{c}{\textbf{N}}     & \multicolumn{1}{c}{\textbf{R}}     & \multicolumn{1}{c}{\textbf{N}}     & \multicolumn{1}{c}{\textbf{R}}     & \multicolumn{1}{c}{\textbf{N}}     & \multicolumn{1}{c}{\textbf{R}}     & \multicolumn{1}{c}{\textbf{N}}     & \multicolumn{1}{c}{\textbf{R}}     \\
\midrule
Llama 2                                                             & 1               & Native          & \multicolumn{1}{c}{22.95}          & \multicolumn{1}{c}{29.21}          & \multicolumn{1}{c}{8.15}           & \multicolumn{1}{c}{11.9}           & \multicolumn{1}{c}{12.12}          & \multicolumn{1}{c}{14.62}          & \multicolumn{1}{c}{15.32}          & \multicolumn{1}{c}{20.22}          & \multicolumn{1}{c}{11.2}           & \multicolumn{1}{c}{14.98}          \\
Llama 2                                                             & 3               & Native          & \multicolumn{1}{c}{23.73}          & \multicolumn{1}{c}{29.91}          & \multicolumn{1}{c}{7.38}           & \multicolumn{1}{c}{11.48}          & \multicolumn{1}{c}{12.4}           & \multicolumn{1}{c}{15.49}          & \multicolumn{1}{c}{16.19}          & \multicolumn{1}{c}{21.4}           & \multicolumn{1}{c}{11.55}          & \multicolumn{1}{c}{15.31}          \\
Llama 2                                                             & 1               & Romanized       & \multicolumn{1}{c}{17.25}          & \multicolumn{1}{c}{18.86}          & \multicolumn{1}{c}{14.63}          & \multicolumn{1}{c}{16.16}          & \multicolumn{1}{c}{12.8}           & \multicolumn{1}{c}{14.87}          & \multicolumn{1}{c}{13.42}          & \multicolumn{1}{c}{15.85}          & \multicolumn{1}{c}{12.88}          & \multicolumn{1}{c}{15.48}          \\
Llama 2                                                             & 3               & Romanized       & \multicolumn{1}{c}{15.01}          & \multicolumn{1}{c}{18.25}          & \multicolumn{1}{c}{10.8}           & \multicolumn{1}{c}{14.79}          & \multicolumn{1}{c}{9.3}            & \multicolumn{1}{c}{11.39}          & \multicolumn{1}{c}{12.09}          & \multicolumn{1}{c}{16.61}          & \multicolumn{1}{c}{12.88}          & \multicolumn{1}{c}{15.53}          \\
\midrule
CPT - N                                                       & 1               & Native          & \multicolumn{1}{c}{32.07}          & \multicolumn{1}{c}{37.89}          & \multicolumn{1}{c}{15.94}          & \multicolumn{1}{c}{18.79}          & \multicolumn{1}{c}{25.81}          & \multicolumn{1}{c}{28.87}          & \multicolumn{1}{c}{29.47}          & \multicolumn{1}{c}{34.24}          & \multicolumn{1}{c}{24.48}          & \multicolumn{1}{c}{28.36}          \\
CPT - N                                                        & 3               & Native          & \multicolumn{1}{c}{32.9}           & \multicolumn{1}{c}{38.51}          & \multicolumn{1}{c}{16.3}           & \multicolumn{1}{c}{19.29}          & \multicolumn{1}{c}{26.2}           & \multicolumn{1}{c}{29.12}          & \multicolumn{1}{c}{29.9}           & \multicolumn{1}{c}{34.75}          & \multicolumn{1}{c}{25.64}          & \multicolumn{1}{c}{29.45}          \\
CPT - R                                                     & 1               & Romanized       & \multicolumn{1}{c}{30.93}          & \multicolumn{1}{c}{33.66}          & \multicolumn{1}{c}{29.73}          & \multicolumn{1}{c}{35.35}          & \multicolumn{1}{c}{30.11}          & \multicolumn{1}{c}{32.71}          & \multicolumn{1}{c}{28.27}          & \multicolumn{1}{c}{33.71}          & \multicolumn{1}{c}{28.72}          & \multicolumn{1}{c}{32.47}          \\
CPT - R                                                     & 3               & Romanized       & \multicolumn{1}{c}{31.51}          & \multicolumn{1}{c}{34.99}          & \multicolumn{1}{c}{30.55}          & \multicolumn{1}{c}{36.8}           & \multicolumn{1}{c}{31.5}           & \multicolumn{1}{c}{35.04}          & \multicolumn{1}{c}{29.42}          & \multicolumn{1}{c}{35.08}          & \multicolumn{1}{c}{30.85}          & \multicolumn{1}{c}{34.86}          \\
\midrule
IFT (Llama 2) - N                                      & 0               & Native          & \multicolumn{1}{c}{43.28}          & \multicolumn{1}{c}{48.43}          & \multicolumn{1}{c}{23.54}          & \multicolumn{1}{c}{25.82}          & \multicolumn{1}{c}{38.99}          & \multicolumn{1}{c}{41.13}          & \multicolumn{1}{c}{40.13}          & \multicolumn{1}{c}{44.72}          & \multicolumn{1}{c}{38.85}          & \multicolumn{1}{c}{42.12}          \\
IFT (CPT - N) - N                                 & 0               & Native          & \multicolumn{1}{c}{44.08}          & \multicolumn{1}{c}{49.12}          & \multicolumn{1}{c}{23.86}          & \multicolumn{1}{c}{25.97}          & \multicolumn{1}{c}{39.3}           & \multicolumn{1}{c}{41.51}          & \multicolumn{1}{c}{40.31}          & \multicolumn{1}{c}{44.8}           & \multicolumn{1}{c}{39.44}          & \multicolumn{1}{c}{42.78}          \\
IFT (Llama 2) - R                                    & 0               & Romanized       & \multicolumn{1}{c}{45.15}          & \multicolumn{1}{c}{55.23}          & \multicolumn{1}{c}{44.8}           & \multicolumn{1}{c}{52.98}          & \multicolumn{1}{c}{49.13}          & \multicolumn{1}{c}{52.79}          & \multicolumn{1}{c}{42.78}          & \multicolumn{1}{c}{49.6}           & \multicolumn{1}{c}{47.71}          & \multicolumn{1}{c}{52.63}          \\
IFT (CPT - R) - R                           & 0               & Romanized       & \multicolumn{1}{c}{\textbf{45.94}}          & \multicolumn{1}{c}{\textbf{56.19}}          & \multicolumn{1}{c}{\textbf{45.37}}          & \multicolumn{1}{c}{\textbf{53.5}}           & \multicolumn{1}{c}{\textbf{50.06}}          & \multicolumn{1}{c}{\textbf{53.71}}          & \multicolumn{1}{c}{\textbf{44}}             & \multicolumn{1}{c}{\textbf{50.73}}          & \multicolumn{1}{c}{\textbf{48.96}} & \multicolumn{1}{c}{\textbf{53.76}} \\
\bottomrule
\end{tabular}
}
\caption{chrF scores for En-XX translation task on FLORES-200 devtest \citep{goyal-etal-2022-flores} across 5 Indic languages in both native (N) and romanized (R) script with different shots.}
\label{langwise-flores-enxx}
\end{table*}

\begin{table*}[]
\small
\centering
\resizebox{0.8\linewidth}{!}{
\begin{tabular}{lcccccccccccc}
\toprule
\multirow{1}{*}{\textbf{Model}} & \multirow{1}{*}{\textbf{Shots}} & \multirow{1}{*}{\textbf{Input text}} & \multicolumn{1}{c}{\textbf{Hindi}} & \multicolumn{1}{c}{\textbf{Gujarati}} &  \multicolumn{1}{c}{\textbf{Tamil}} & \multicolumn{1}{c}{\textbf{Marathi}} & \multicolumn{1}{c}{\textbf{Malayalam}} \\
\midrule
Llama 2                                                             & 1               & Native                    & 40.49         & 13.63         & 16.06          & 24.18          & 17.93          \\
Llama 2                                                             & 3               & Native                     & 40.38         & 17.58         & 17.22          & 24.15          & 17.78          \\
Llama 2                                                             & 1               & Romanized                  & 29.24         & 22.62         & 17.06          & 22.56          & 21.1          \\
Llama 2                                                             & 3               & Romanized                   & 28.91         & 23.28         & 19.86          & 23.6           & 21.96          \\
\midrule
CPT - N                                                       & 1               & Native                    & 51.37         & 24.77         & 33.2           & 47.04          & 36.34          \\
CPT - N                                                        & 3               & Native                     & 50.31         & 25.73         & 28.56          & 43.37          & 37.4          \\
CPT - R                                                     & 1               & Romanized                 & 45.43         & 45.08         & 38.34          & 44.28          & 39.52          \\
CPT - R                                                     & 3               & Romanized                  & 45.4          & 44.82         & 33.72          & 43.56          & 40.71          \\
\midrule
IFT (Llama 2) - N                                      & 0               & Native                    & 51.7          & 47.26         & 42.27          & 48.8           & 47.83          \\
IFT (CPT - N) - N                                 & 0               & Native                     & 53.67         & 49.51         & 44.44          & \textbf{51.29} & 50          \\
IFT (Llama 2) - R                                    & 0               & Romanized                 & 51.84         & 50.17         & 41.67          & 48.51          & 47.14          \\
IFT (CPT - R) - R                           & 0               & Romanized                 & \textbf{53.89}         & \textbf{53.3} & \textbf{45.09}          & 51.06          & \textbf{50.43} \\
\bottomrule
\end{tabular}
}
\caption{chrF scores for XX-En translation task on FLORES-200 devtest \citep{goyal-etal-2022-flores} across 5 Indic languages in both native (N) and romanized (R) script with different shots.}
\label{langwise-flores-xxen}
\end{table*}

\begin{table*}[]
\small
\centering
\resizebox{0.9\linewidth}{!}{
\begin{tabular}{lcccccccccc}
\toprule
\multirow{2}{*}{\textbf{Model}} & \multirow{2}{*}{\textbf{Shots}} & \multirow{2}{*}{\textbf{Input text}} & \multicolumn{2}{c}{\textbf{Hindi}} & \multicolumn{2}{c}{\textbf{Gujarati}} &  \multicolumn{2}{c}{\textbf{Tamil}} & \multicolumn{2}{c}{\textbf{Marathi}}  \\
\cmidrule{4-11}
& & & \multicolumn{1}{c}{\textbf{N}}     & \multicolumn{1}{c}{\textbf{R}}     & \multicolumn{1}{c}{\textbf{N}}     & \multicolumn{1}{c}{\textbf{R}}     & \multicolumn{1}{c}{\textbf{N}}     & \multicolumn{1}{c}{\textbf{R}}     & \multicolumn{1}{c}{\textbf{N}}     & \multicolumn{1}{c}{\textbf{R}}          \\
\midrule
Llama 2                                                             & 1               & Native         & 13.32          & 9.59           & 3.2           & 2.79           & 4.13           & 3.47           & 6.85           & 6.11           \\
Llama 2                                                             & 1               & Romanized    & 13.43          & 15.96          & 8.95          & 7.65           & 9.05           & 6.69           & 9.22           & 7.45             \\
\midrule
CPT - N                                                       & 1               & Native         & \textbf{16.32} & 12.18          & 5.18          & 4.58           & 4.3            & 3.91           & 4.55           & 4.61          \\
CPT - R                                                     & 1               & Romanized       & 13.05          & 14.77          & \textbf{13.8} & \textbf{12.98} & 10.86          & \textbf{9.68}           & \textbf{12.06}          & \textbf{11.34}          \\
\midrule
IFT (Llama 2) - N                                      & 0               & Native     & 11.71          & 8.66           & 4.38          & 3.18           & 6.62           & 5.62           & 7.41           & 6.61               \\
IFT (CPT - N) - N                                 & 0               & Native      & 12.56          & 9.21           & 4.3           & 3.52           & 6.51           & 5.66           & 7.71           & 6.64              \\
IFT (Llama 2) - R                                    & 0               & Romanized      & 14.73          & 16.87          & 11.84         & 10.56          & 10.83          & 8.59           & 11             & 9.53           \\
IFT (CPT - R) - R                           & 0               & Romanized   & 15.05          & \textbf{17.92}          & 12.36         & 11.16          & \textbf{11.28}          & 9.32           & 11.56          & 9.87     \\
\bottomrule
\end{tabular}
}
\caption{RougeL scores on XLSum \citep{hasan-etal-2021-xl} across 5 Indic languages in both native (N) and romanized (R) script with different shots.}
\label{langwise-xlsum}
\end{table*}

\begin{table*}[]
\small
\centering
\resizebox{\linewidth}{!}{
\begin{tabular}{lcccccccccccc}
\toprule
\multirow{2}{*}{\textbf{Model}} & \multirow{2}{*}{\textbf{n-shot}} & \multirow{2}{*}{\textbf{Input text}} & \multicolumn{2}{c}{\textbf{Hindi}} & \multicolumn{2}{c}{\textbf{Gujarati}} &  \multicolumn{2}{c}{\textbf{Tamil}} & \multicolumn{2}{c}{\textbf{Marathi}} & \multicolumn{2}{c}{\textbf{Malayalam}} \\
\cmidrule{4-13}
& & & \multicolumn{1}{c}{\textbf{N}}     & \multicolumn{1}{c}{\textbf{R}}     & \multicolumn{1}{c}{\textbf{N}}     & \multicolumn{1}{c}{\textbf{R}}     & \multicolumn{1}{c}{\textbf{N}}     & \multicolumn{1}{c}{\textbf{R}}     & \multicolumn{1}{c}{\textbf{N}}     & \multicolumn{1}{c}{\textbf{R}}        & \multicolumn{1}{c}{\textbf{N}}     & \multicolumn{1}{c}{\textbf{R}}      \\
\midrule
Llama 2                                                             & 1               & Native    & 20.91 & 15.01 & 6.17  & 5.55  & 19.14 & 17.8  & 9.26  & 8.45  & 12.84 & 11.45      \\
Llama 2                                                             & 1               & Romanized   & 17.58 & 20.55 & 12.18 & 13.72 & 23.5  & 26.16 & 10.55 & 11.32 & 13.97 & 18.83          \\
\midrule
CPT - N                                                       & 1               & Native  & \textbf{22.74} & 16.52 & 9.07  & 7.97  & 23.59 & 22.79 & \textbf{19.21} & \textbf{18.72} & 15.57 & 14.74        \\
CPT - R                                                     & 1               & Romanized  & 20.19 & \textbf{24.44} & 12.83 & 14.34 & \textbf{29.13} & \textbf{32.03} & 15.14 & 17.39 & \textbf{17.29} & \textbf{22.38}        \\
\midrule
IFT (Llama 2) - N                                      & 0               & Native   & 18.38 & 13.43 & 6.36  & 5.64  & 15.88 & 15.35 & 9.96  & 9.84  & 10.51 & 9.85           \\
IFT (CPT - N) - N                                 & 0               & Native     & 19.47 & 14.24 & 6.65  & 5.98  & 15.14 & 14.75 & 10.77 & 10.7  & 11.04 & 10.68         \\
IFT (Llama 2) - R                                    & 0               & Romanized     & 17.76 & 21.46 & 12.88 & 15.43 & 20.89 & 22.27 & 11.51 & 12.62 & 13.45 & 17.45      \\
IFT (CPT - R) - R                           & 0               & Romanized   & 18.74 & 21.91 & \textbf{13.04} & \textbf{15.64} & 21.61 & 23.82 & 12.2  & 13.59 & 14.56 & 19.27  \\
\bottomrule
\end{tabular}
}
\caption{RougeL scores on IndicHeadline \citep{kumar-etal-2022-indicnlg} across 5 Indic languages in both native (N) and romanized (R) script with different shots.}
\label{langwise-indicheadline}
\end{table*}

\begin{table*}[]
\small
\centering
\resizebox{0.85\linewidth}{!}{
\begin{tabular}{lcccccccccccc}
\toprule
\multirow{1}{*}{\textbf{Model}} & \multirow{1}{*}{\textbf{n-shot}} & \multirow{1}{*}{\textbf{Input text}} & \multicolumn{1}{c}{\textbf{Hindi}} & \multicolumn{1}{c}{\textbf{Gujarati}} &  \multicolumn{1}{c}{\textbf{Tamil}} & \multicolumn{1}{c}{\textbf{Marathi}} & \multicolumn{1}{c}{\textbf{Malayalam}} \\
\midrule
Llama 2                                                             & 1               & Native     & 28.67 & 25.59 & 26.06 & 27.58 & 26.44     \\
Llama 2                                                             & 3               & Native    & 29.11 & 26.57 & 26.75 & 27.85 & 26.86      \\
Llama 2                                                             & 1               & Romanized    & 27.52 & 26.57 & 26.34 & 27.14 & 27.11      \\
Llama 2                                                             & 3               & Romanized    & 28.38 & 27.62 & 26.65 & 27.24 & 27.52      \\
\midrule
CPT - N                                                       & 1               & Native    & 29.61 & 27.34 & 28.63 & 29.58 & 28.35      \\
CPT - N                                                        & 3               & Native   & 30.08 & 27.62 & 28.66 & 29.91 & 28.61       \\
CPT - R                                                     & 1               & Romanized    & 30.8  & 30.72 & \textbf{30.38} & 31.33 & 29.7      \\
CPT - R                                                     & 3               & Romanized    & \textbf{31.55} & \textbf{31.05} & 30.18 & \textbf{32.15} & \textbf{30.76}      \\
\midrule
IFT (Llama 2) - N                                      & 0               & Native    & 24.47 & 24.08 & 24.09 & 23.3  & 24.36      \\
IFT (CPT - N) - N                                 & 0               & Native    & 24.72 & 24.05 & 24.78 & 24.62 & 24.81      \\
IFT (Llama 2) - R                                    & 0               & Romanized     & 24.75 & 25.96 & 26.03 & 25.54 & 26.48     \\
IFT (CPT - R) - R                           & 0               & Romanized   & 25.58 & 26.4  & 26.73 & 26.1  & 26.57  \\
\bottomrule
\end{tabular}
}
\caption{Accuracy on translated MMLU \citep{hendrycks2021measuring} across 5 Indic languages in both native (N) and romanized (R) script with different shots.}
\label{langwise-mmlu}
\end{table*}

\begin{table*}[]
\small
\centering
\resizebox{0.85\linewidth}{!}{
\begin{tabular}{lcccccccccccc}
\toprule
\multirow{1}{*}{\textbf{Model}} & \multirow{1}{*}{\textbf{n-shot}} & \multirow{1}{*}{\textbf{Input text}} & \multicolumn{1}{c}{\textbf{Hindi}} & \multicolumn{1}{c}{\textbf{Gujarati}} &  \multicolumn{1}{c}{\textbf{Tamil}} & \multicolumn{1}{c}{\textbf{Marathi}} & \multicolumn{1}{c}{\textbf{Malayalam}} \\
\midrule
Llama 2                    & 1               & Native   & 57.58 & 58.78 & 56.88 & 61.07 & 55.2       \\
Llama 2                    & 3               & Native   & 45.08 & 57.58 & 39.3  & 40.92 & 39.39       \\
Llama 2                    & 1               & Romanized   & 62.23 & 60.03 & 58.72 & 60.61 & 62.14       \\
Llama 2                    & 3               & Romanized    & 61.93 & \textbf{61.77} & 60.09 & 58.9  & \textbf{62.48}      \\
\midrule
CPT - N                     & 1               & Native    & 60.7  & 56.21 & 57.71 & 61.41 & 59.45      \\
CPT - N                      & 3               & Native     & 50.67 & 60.64 & 45.78 & 39.36 & 40.95     \\
CPT - R                       & 1               & Romanized    & 62.17 & 61.5  & \textbf{61.44} & \textbf{61.93} & 62.23      \\
CPT - R                       & 3               & Romanized    & \textbf{64.77} & 60.92 & 55.41 & 47.86 & 56.54      \\
\midrule
IFT (Llama 2) - N             & 0               & Native     & 60.12 & 61.65 & 60.06 & 61.01 & 60.18     \\
IFT (CPT - N) - N             & 0               & Native     & 60.34 & 61.5  & 58.32 & 60.43 & 59.6     \\
IFT (Llama 2) - R             & 0               & Romanized    & 61.62 & 56.91 & 58.01 & 58.69 & 47.68      \\
IFT (CPT - R) - R             & 0               & Romanized & 60.7  & 45.02 & 38.01 & 53.24 & 37.77 \\
\bottomrule
\end{tabular}
}
\caption{Accuracy on translated BoolQ \citep{clark2019boolq} across 5 Indic languages in both native (N) and romanized (R) script with different shots.}
\label{langwise-boolq}
\end{table*}

\begin{table*}[]
\small
\centering
\resizebox{0.85\linewidth}{!}{
\begin{tabular}{lcccccccccccc}
\toprule
\multirow{1}{*}{\textbf{Model}} & \multirow{1}{*}{\textbf{n-shot}} & \multirow{1}{*}{\textbf{Input text}} & \multicolumn{1}{c}{\textbf{Hindi}} & \multicolumn{1}{c}{\textbf{Gujarati}} &  \multicolumn{1}{c}{\textbf{Tamil}} & \multicolumn{1}{c}{\textbf{Marathi}} & \multicolumn{1}{c}{\textbf{Malayalam}} \\
\midrule
Llama 2               & 1               & Native    & 27.27 & 25.67 & 25.63 & 25.13 & 25.42      \\
Llama 2               & 3               & Native     & 27.36 & 26.09 & 26.18 & 26.6  & 25.51     \\
Llama 2               & 1               & Romanized   & 27.23 & 25.29 & 26.05 & 26.98 & 26.35       \\
Llama 2               & 3               & Romanized    & 27.27 & \textbf{26.64} & 25.29 & 26.85 & 26.14       \\
\midrule
CPT - N               & 1               & Native    & \textbf{30.98} & 26.22 & \textbf{28.87} & \textbf{30.56} & 27.23      \\
CPT - N               & 3               & Native     & 30.68 & 26.43 & 27.4  & 29.12 & \textbf{27.61}     \\
CPT - R               & 1               & Romanized   & 27.23 & 25.88 & 26.26 & 26.89 & 27.19       \\
CPT - R               & 3               & Romanized    & 27.61 & 26.56 & 26.39 & 27.74 & 27.23      \\
\midrule
IFT (Llama 2) - N       & 0               & Native     & 25.42 & 25.88 & 25.25 & 25.42 & 26.09     \\
IFT (CPT - N) - N       & 0               & Native     & 26.35 & 25.84 & 25.67 & 25.93 & 26.26     \\
IFT (Llama 2) - R       & 0               & Romanized     & 25.08 & 25.72 & 25.04 & 25.34 & 25.76     \\
IFT (CPT - R) - R       & 0               & Romanized   & 25.97 & 25.34 & 24.24 & 25.42 & 25.72 \\
\bottomrule
\end{tabular}
}
\caption{Accuracy on translated ARC-Easy \citep{allenai:arc} across 5 Indic languages in both native (N) and romanized (R) script with different shots.}
\label{langwise-arc-easy}
\end{table*}

\begin{table*}[]
\small
\centering
\resizebox{0.85\linewidth}{!}{
\begin{tabular}{lcccccccccccc}
\toprule
\multirow{1}{*}{\textbf{Model}} & \multirow{1}{*}{\textbf{n-shot}} & \multirow{1}{*}{\textbf{Input text}} & \multicolumn{1}{c}{\textbf{Hindi}} & \multicolumn{1}{c}{\textbf{Gujarati}} &  \multicolumn{1}{c}{\textbf{Tamil}} & \multicolumn{1}{c}{\textbf{Marathi}} & \multicolumn{1}{c}{\textbf{Malayalam}} \\
\midrule
Llama 2                       & 1               & Native    & 26.45 & 27.05 & 26.71 & 26.71 & 25.77      \\
Llama 2                       & 3               & Native    & 27.22 & 26.19 & 25.17 & 25.6  & 24.32      \\
Llama 2                       & 1               & Romanized    & 24.4  & 25.68 & 24.32 & 24.32 & 24.32      \\
Llama 2                       & 3               & Romanized    & 24.83 & 27.47 & 25.09 & 25.09 & 25.34      \\
\midrule
CPT - N                       & 1               & Native    & 27.65 & 26.62 & \textbf{27.3}  & 27.3  & \textbf{28.92}      \\
CPT - N                       & 3               & Native    & \textbf{29.69} & 25.85 & 26.96 & 26.96 & 26.54      \\
CPT - R                       & 1               & Romanized    & 25.34 & 26.79 & 24.57 & \textbf{28.07} & 27.22      \\
CPT - R                       & 3               & Romanized     & 24.49 & \textbf{27.56} & 25.94 & 27.99 & 27.3     \\
\midrule
IFT (Llama 2) - N             & 0               & Native   & 22.27 & 22.78 & 23.46 & 21.93 & 23.29       \\
IFT (CPT - N) - N             & 0               & Native   & 25.85 & 23.21 & 24.57 & 24.06 & 23.29       \\
IFT (Llama 2) - R             & 0               & Romanized    & 21.67 & 22.87 & 23.04 & 22.87 & 21.67      \\
IFT (CPT - R) - R             & 0               & Romanized    & 22.78 & 22.95 & 22.7  & 22.35 & 23.21     \\
\bottomrule
\end{tabular}
}
\caption{Accuracy on translated ARC-Challenge \citep{allenai:arc} across 5 Indic languages in both native (N) and romanized (R) script with different shots.}
\label{langwise-arc-challenge}
\end{table*}

\begin{table*}[]
\small
\centering
\resizebox{0.85\linewidth}{!}{
\begin{tabular}{lcccccccccccc}
\toprule
\multirow{1}{*}{\textbf{Model}} & \multirow{1}{*}{\textbf{n-shot}} & \multirow{1}{*}{\textbf{Input text}} & \multicolumn{1}{c}{\textbf{Hindi}} & \multicolumn{1}{c}{\textbf{Gujarati}} &  \multicolumn{1}{c}{\textbf{Tamil}} & \multicolumn{1}{c}{\textbf{Marathi}} & \multicolumn{1}{c}{\textbf{Malayalam}} \\
\midrule
Llama 2                     & 1               & Native     & 23.99 & 19    & 20.55 & 21.45 & 20.06     \\
Llama 2                     & 3               & Native     & 23.75 & 18.75 & 20.22 & 20.22 & 19.32     \\
Llama 2                     & 1               & Romanized    & 21.54 & 20.48 & 19.49 & 19.74 & 19.08      \\
Llama 2                     & 3               & Romanized    & 21.54 & 19.57 & 19.98 & 19.82 & 19.98      \\
\midrule
CPT - N                     & 1               & Native    & 28.09 & 23.75 & 28.5  & 26.94 & 25.47      \\
CPT - N                     & 3               & Native    & \textbf{31.12} & 22.44 & \textbf{28.99} & \textbf{28.09} & 26.45      \\
CPT - R                     & 1               & Romanized    & 23.75 & 27.19 & 25.3  & 25.22 & 26.28      \\
CPT - R                     & 3               & Romanized     & 25.71 & \textbf{30.22}& 28    & 27.1  & \textbf{29.89}     \\
\midrule
IFT (Llama 2) - N           & 0               & Native    & 20.64 & 19.25 & 20.07 & 20.97 & 20.31      \\
IFT (CPT - N) - N           & 0               & Native    & 23.34 & 23.18 & 23.59 & 25.14 & 25.55      \\
IFT (Llama 2) - R           & 0               & Romanized     & 20.23 & 21.54 & 21.29 & 22.93 & 22.44     \\
IFT (CPT - R) - R           & 0               & Romanized    & 23.91 & 26.7  & 25.55 & 26.04 & 27.52     \\
\bottomrule
\end{tabular}
}
\caption{Accuracy on translated CommonsenseQA \citep{talmor-etal-2019-commonsenseqa} across 5 Indic languages in both native (N) and romanized (R) script with different shots.}
\label{langwise-commonsenseqa}
\end{table*}

\begin{table*}[]
\small
\centering
\resizebox{0.85\linewidth}{!}{
\begin{tabular}{lcccccccccccc}
\toprule
\multirow{1}{*}{\textbf{Model}} & \multirow{1}{*}{\textbf{n-shot}} & \multirow{1}{*}{\textbf{Input text}} & \multicolumn{1}{c}{\textbf{Hindi}} & \multicolumn{1}{c}{\textbf{Gujarati}} &  \multicolumn{1}{c}{\textbf{Tamil}} & \multicolumn{1}{c}{\textbf{Marathi}} & \multicolumn{1}{c}{\textbf{Malayalam}} \\
\midrule
Llama 2                        & 1               & Native     & 53.2 & 35.1 & 46   & 50.3 & 40.4     \\
Llama 2                        & 3               & Native     & 77.8 & 28.4 & 34.5 & 49.2 & 31.7     \\
Llama 2                        & 1               & Romanized     & 59.1 & 44.9 & 42.4 & 53.5 & 40.9     \\
Llama 2                        & 3               & Romanized   & 67.5 & 38.9 & 33.8 & 44.3 & 34       \\
\midrule
CPT - N                        & 1               & Native     & 64.9 & 68   & 77.8 & 75.1 & 67.1     \\
CPT - N                        & 3               & Native     & 89.7 & 86.6 & 87.8 & 89.6 & 82.9     \\
CPT - R                        & 1               & Romanized     & 91.2 & 86.6 & 88.4 & 90   & 86.1     \\
CPT - R                        & 3               & Romanized     & \textbf{91.9} & \textbf{93.3} & \textbf{93.3} & \textbf{93.1} & \textbf{92.5}     \\
\midrule
IFT (Llama 2) - N             & 0               & Native    & 90.4 & 88.1 & 87.6 & 89.5 & 89      \\
IFT (CPT - N) - N              & 0               & Native     & 87.2 & 88.7 & 89.6 & 90   & 88.9     \\
IFT (Llama 2) - R              & 0               & Romanized     & 75.3 & 68.8 & 71.2 & 77.6 & 77.3     \\
IFT (CPT - R) - R              & 0               & Romanized     & 83.7 & 73.5 & 87.7 & 81.9 & 85.4    \\
\bottomrule
\end{tabular}
}
\caption{F1 score on IndicSentiment \citep{doddapaneni-etal-2023-towards} across 5 Indic languages in both native (N) and romanized (R) script with different shots.}
\label{langwise-indicsentiment}
\end{table*}

\begin{table*}[]
\small
\centering
\resizebox{0.85\linewidth}{!}{
\begin{tabular}{lcccccccccccc}
\toprule
\multirow{1}{*}{\textbf{Model}} & \multirow{1}{*}{\textbf{n-shot}} & \multirow{1}{*}{\textbf{Input text}} & \multicolumn{1}{c}{\textbf{Hindi}} & \multicolumn{1}{c}{\textbf{Gujarati}} &  \multicolumn{1}{c}{\textbf{Tamil}} & \multicolumn{1}{c}{\textbf{Marathi}} & \multicolumn{1}{c}{\textbf{Malayalam}} \\
\midrule
Llama 2                    & 1               & Native     & 41.1 & 41.1 & 1.6  & 41.1 & 1.6     \\
Llama 2                    & 3               & Native     & 52.1 & 56.7 & 7.5  & 34.7 & 1.6     \\
Llama 2                    & 1               & Romanized    & 39.2 & 36.2 & 1.6  & 26.6 & 1.6      \\
Llama 2                    & 3               & Romanized     & 23.6 & 16.1 & 3.1  & 5.9  & 0     \\
\midrule
CPT - N                    & 1               & Native     & 40.5 & 40.7 & 1.6  & 38.4 & 1.6     \\
CPT - N                    & 3               & Native     & \textbf{61}   & \textbf{65.7} & 11.6 & \textbf{60}   & 11.7     \\
CPT - R                    & 1               & Romanized    & 38.6 & 37.4 & 0    & 31.8 & 1.6      \\
CPT - R                    & 3               & Romanized     & 30.3 & 42.7 & 6.6  & 32.3 & 0.8     \\
\midrule
IFT (Llama 2) - N         & 0               & Native    & 5.2  & 1.7  & 44.1 & 1.8  & 7.4      \\
IFT (CPT - N) - N         & 0               & Native    & 51.3 & 33.2 & 47.6 & 35.4 & 56.3      \\
IFT (Llama 2) - R         & 0               & Romanized      & 18.9 & 58.4 & 18.8 & 35.9 & 63.6    \\
IFT (CPT - R) - R         & 0               & Romanized      & 33.2 & 62.3 & \textbf{63.4} & 47.1 & \textbf{66.7}     \\
\bottomrule
\end{tabular}
}
\caption{F1 score on IndicCOPA \citep{doddapaneni-etal-2023-towards} across 5 Indic languages in both native (N) and romanized (R) script with different shots.}
\label{langwise-indiccopa}
\end{table*}

\begin{table*}[]
\small
\centering
\resizebox{0.85\linewidth}{!}{
\begin{tabular}{lcccccccccccc}
\toprule
\multirow{1}{*}{\textbf{Model}} & \multirow{1}{*}{\textbf{n-shot}} & \multirow{1}{*}{\textbf{Input text}} & \multicolumn{1}{c}{\textbf{Hindi}} & \multicolumn{1}{c}{\textbf{Gujarati}} &  \multicolumn{1}{c}{\textbf{Tamil}} & \multicolumn{1}{c}{\textbf{Marathi}} & \multicolumn{1}{c}{\textbf{Malayalam}} \\
\midrule
Llama 2                     & 1               & Native     & 17.6 & 16.7 & 16.8 & 17.1 & 16.7     \\
Llama 2                     & 3               & Native     & 27.4 & 25.1 & 17.4 & 18.7 & 16.9     \\
Llama 2                     & 1               & Romanized     & 27.5 & 28.5 & 26.5 & 26.4 & 27.3     \\
Llama 2                     & 3               & Romanized     & 31.4 & 31.6 & 30.5 & 31.2 & 31.1     \\
\midrule
CPT - N                     & 1               & Native    & 19.9 & 19.3 & 19.4 & 17.7 & 17.1      \\
CPT - N                     & 3               & Native    & \textbf{44.6} & 30.1 & 32.8 & \textbf{45.2} & 26.1      \\
CPT - R                     & 1               & Romanized     & 34.1 & 37   & 36.6 & 37.3 & 39     \\
CPT - R                     & 3               & Romanized     & 38.6 & 37.3 & 37.5 & 37.3 & 35.7     \\
\midrule
IFT (Llama 2) - N            & 0               & Native    & 36.3 & 37.9 & \textbf{43.1} & 42.9 & 38.7      \\
IFT (CPT - N) - N            & 0               & Native    & 42.2 & 41.2 & \textbf{43.1} & 42.5 & \textbf{42.5}      \\
IFT (Llama 2) - R            & 0               & Romanized     & 24   & 22.5 & 23   & 22.1 & 28.6     \\
IFT (CPT - R) - R            & 0               & Romanized     & 38.5 & \textbf{38.9} & 37.2 & 38.1 & 39.2     \\
\bottomrule
\end{tabular}
}
\caption{F1 score on IndicXNLI \citep{doddapaneni-etal-2023-towards} across 5 Indic languages in both native (N) and romanized (R) script with different shots.}
\label{langwise-indicxnli}
\end{table*}

\begin{table*}[]
\small
\centering
\resizebox{\linewidth}{!}{
\begin{tabular}{lcccccccccccc}
\toprule
\multirow{2}{*}{\textbf{Model}} & \multirow{2}{*}{\textbf{n-shot}} & \multirow{2}{*}{\textbf{Input text}} & \multicolumn{2}{c}{\textbf{Hindi}} & \multicolumn{2}{c}{\textbf{Gujarati}} &  \multicolumn{2}{c}{\textbf{Tamil}} & \multicolumn{2}{c}{\textbf{Marathi}} & \multicolumn{2}{c}{\textbf{Malayalam}} \\
\cmidrule{4-13}
& & & \multicolumn{1}{c}{\textbf{N}}     & \multicolumn{1}{c}{\textbf{R}}     & \multicolumn{1}{c}{\textbf{N}}     & \multicolumn{1}{c}{\textbf{R}}     & \multicolumn{1}{c}{\textbf{N}}     & \multicolumn{1}{c}{\textbf{R}}     & \multicolumn{1}{c}{\textbf{N}}     & \multicolumn{1}{c}{\textbf{R}}        & \multicolumn{1}{c}{\textbf{N}}     & \multicolumn{1}{c}{\textbf{R}}      \\
\midrule
Llama 2                                                             & 1               & Native     & 21.02 & 24.63 & 3.26  & 3.39  & 8.75  & 8.91  & 10.08 & 12.22 & 5.73  & 5.73     \\
Llama 2                                                             & 1               & Romanized     & 29.11 & 32.8  & 19.95 & 21.97 & 11.34 & 13.32 & 21.11 & 21.6  & 17.18 & 20.42        \\
\midrule
CPT - N                                                       & 1               & Native    & 28.98 & 30.36 & 17.01 & 17.76 & 21.01 & 21.04 & 28.76 & 28.96 & 23.37 & 23.43      \\
CPT - R                                                     & 1               & Romanized    & \textbf{29.58} & 32.8  & \textbf{27.01} & \textbf{28.85} & \textbf{24.69} & \textbf{27.11} & \textbf{30.01} & \textbf{30.53} & \textbf{24.95} & \textbf{27.51}      \\
\midrule
IFT (Llama 2) - N                                      & 0               & Native     & 19.75 & 22.79 & 11.66 & 14.3  & 17.56 & 17.81 & 18.39 & 21.96 & 11.15 & 11.4         \\
IFT (CPT - N) - N                                 & 0               & Native      & 21.85 & 25.68 & 13.54 & 16.21 & 18.49 & 18.68 & 18.02 & 21.43 & 12.07 & 12.49        \\
IFT (Llama 2) - R                                    & 0               & Romanized      & 23.54 & 33.79 & 19.15 & 24.3  & 16.44 & 25.54 & 21.26 & 25.79 & 13.2  & 20.31     \\
IFT (CPT - R) - R                           & 0               & Romanized   & 24.44 & \textbf{33.81} & 21.41 & 26.9  & 17.65 & 25.27 & 23.64 & 28.41 & 14.53 & 22.01  \\
\bottomrule
\end{tabular}
}
\caption{F1 score on IndicQA (with context) \citep{doddapaneni-etal-2023-towards} across 5 Indic languages in both native (N) and romanized (R) script with different shots.}
\label{langwise-indicqa-with-context}
\end{table*}

\begin{table*}[]
\small
\centering
\resizebox{\linewidth}{!}{
\begin{tabular}{lcccccccccccc}
\toprule
\multirow{2}{*}{\textbf{Model}} & \multirow{2}{*}{\textbf{n-shot}} & \multirow{2}{*}{\textbf{Input text}} & \multicolumn{2}{c}{\textbf{Hindi}} & \multicolumn{2}{c}{\textbf{Gujarati}} &  \multicolumn{2}{c}{\textbf{Tamil}} & \multicolumn{2}{c}{\textbf{Marathi}} & \multicolumn{2}{c}{\textbf{Malayalam}} \\
\cmidrule{4-13}
& & & \multicolumn{1}{c}{\textbf{N}}     & \multicolumn{1}{c}{\textbf{R}}     & \multicolumn{1}{c}{\textbf{N}}     & \multicolumn{1}{c}{\textbf{R}}     & \multicolumn{1}{c}{\textbf{N}}     & \multicolumn{1}{c}{\textbf{R}}     & \multicolumn{1}{c}{\textbf{N}}     & \multicolumn{1}{c}{\textbf{R}}        & \multicolumn{1}{c}{\textbf{N}}     & \multicolumn{1}{c}{\textbf{R}}      \\
\midrule
Llama 2                                                             & 1               & Native    & 4.97  & 6.69  & 0.2   & 0.22  & 1.02  & 1.09  & 0.9   & 1.16  & 0.31  & 0.36      \\
Llama 2                                                             & 1               & Romanized   & 6.96  & 7.33  & \textbf{16.77} & 16.82 & 12.26 & 12.26 & 16.45 & 16.38 & 12.34 & 12.77          \\
\midrule
CPT - N                                                       & 1               & Native    & 7.44  & 8.67  & 2.73  & 3.15  & 4.66  & 4.64  & 7.41  & 7.82  & 5.55  & 5.67      \\
CPT - R                                                     & 1               & Romanized    & \textbf{13.85} & \textbf{15.19} & 13.46 & \textbf{13.41} & \textbf{16.66} & \textbf{17.13} & \textbf{26.75} & \textbf{26.82} & \textbf{17.38} & \textbf{17.93}      \\
\midrule
IFT (Llama 2) - N                                      & 0               & Native     & 2.36  & 2.64  & 0.66  & 0.72  & 1.48  & 1.53  & 1.18  & 1.28  & 0.89  & 0.9         \\
IFT (CPT - N) - N                                 & 0               & Native     & 2.73  & 3.03  & 0.85  & 0.95  & 1.15  & 1.16  & 1.27  & 1.47  & 0.82  & 0.9         \\
IFT (Llama 2) - R                                    & 0               & Romanized    & 2.21  & 2.93  & 0.65  & 0.76  & 1.37  & 1.42  & 1.75  & 1.75  & 1.01  & 1.33       \\
IFT (CPT - R) - R                           & 0               & Romanized   & 3.32  & 3.69  & 1.03  & 0.96  & 1.4   & 1.67  & 1.81  & 1.9   & 1.05  & 1.21  \\
\bottomrule
\end{tabular}
}
\caption{F1 score on IndicQA (without context) \citep{doddapaneni-etal-2023-towards} across 5 Indic languages in both native (N) and romanized (R) script with different shots.}
\label{langwise-indicqa-without-context}
\end{table*}

\begin{table*}[t]
\centering
\small
\resizebox{0.8\linewidth}{!}{
\begin{tabular}{lccccccc}
\toprule
\textbf{Model} & \textbf{Shots} & \textbf{Input text} & \textbf{Hindi} & \textbf{Gujarati} & \textbf{Tamil} & \textbf{Marathi} & \textbf{Malayalam}\\
\midrule
IFT (BLOOM) - N & 0 & Native & 51.03 & 48.39 & \textbf{52.1} & 45.63 & \textbf{49.94} \\
IFT (BLOOM) - R	& 0 & Romanized & \textbf{53.74} & \textbf{50.95} & 50.09 & \textbf{48.06} & 48.93 \\

\bottomrule
\end{tabular}
}
\caption{chrF scores for En-XX translation task on FLORES-200 devtest \citep{goyal-etal-2022-flores} across 5 Indic languages in both native and romanized script.}
\label{tab:bloom-mt}
\end{table*}

\begin{table*}[t]
\centering
\small
\resizebox{0.8\linewidth}{!}{
\begin{tabular}{lccccccc}
\toprule
\textbf{Model} & \textbf{Shots} & \textbf{Input text} & \textbf{Hindi} & \textbf{Gujarati} & \textbf{Tamil} & \textbf{Marathi} & \textbf{Malayalam}\\
\midrule
IFT (BLOOM) - N & 0-shot & Native & 7.44 & 4.36 & 7.66 & 0.8 & 3.7 \\
IFT (BLOOM) - R & 0-shot & Romanized & \textbf{20.96} & \textbf{15.87} & \textbf{25.4}	& \textbf{13.46} & \textbf{19.42} \\

\bottomrule
\end{tabular}
}
\caption{RougeL scores on IndicHeadline \citep{kumar-etal-2022-indicnlg} across 5 Indic languages in both native and romanized script.}
\label{tab:bloom-indicheadline}
\end{table*}

\begin{table*}[t]
\centering
\small
\resizebox{0.75\linewidth}{!}{
\begin{tabular}{lcccccc}
\toprule
\textbf{Model} & \textbf{Shots} & \textbf{Input text} & \textbf{Hindi} & \textbf{Gujarati} & \textbf{Tamil} & \textbf{Marathi} \\
\midrule
IFT (BLOOM) - N & 0-shot & Native & 1.73 & 2.84 & 2.56 & 2.73  \\
IFT (BLOOM) - R & 0-shot & Romanized & \textbf{17.16} & \textbf{10.74} & \textbf{8.63}	& \textbf{9.36}  \\

\bottomrule
\end{tabular}
}
\caption{RougeL scores on XLSum \citep{hasan-etal-2021-xl} across 5 Indic languages in both native and romanized script.}
\label{tab:bloom-xlsm}
\end{table*}

\end{document}